\RequirePackage{fix-cm}
\documentclass[11pt]{digital_agent_hub}
\DeclareFontShape{T1}{productsans}{m}{it}{<-> s * [1] \dahfontdir/productsansregular}{}
\DeclareFontShape{T1}{productsans}{m}{sl}{<-> s * [1] \dahfontdir/productsansregular}{}
\DeclareFontShape{T1}{productsans}{m}{sc}{<-> s * [1] \dahfontdir/productsansregular}{}
\DeclareFontShape{T1}{productsans}{b}{it}{<-> s * [1] \dahfontdir/productsansbold}{}
\DeclareFontShape{T1}{productsans}{b}{sc}{<-> s * [1] \dahfontdir/productsansbold}{}
\DeclareFontShape{T1}{productsans}{bx}{it}{<-> s * [1] \dahfontdir/productsansbold}{}
\DeclareFontShape{T1}{productsans}{bx}{sc}{<-> s * [1] \dahfontdir/productsansbold}{}
\titleformat{\paragraph}[runin]{\normalfont\normalsize\bfseries}{}{0pt}{}
\titlespacing*{\paragraph}{0pt}{0.6\baselineskip}{0.6em}

\usepackage{amsmath}

\usepackage{amsmath,amsfonts,bm}

\def\eqref#1{equation~\ref{#1}}

\def\1{\bm{1}}

\def\mK{{\bm{K}}}

\def\mR{{\bm{R}}}

\DeclareMathAlphabet{\mathsfit}{\encodingdefault}{\sfdefault}{m}{sl}
\SetMathAlphabet{\mathsfit}{bold}{\encodingdefault}{\sfdefault}{bx}{n}

\usepackage{booktabs}
\usepackage{graphicx}
\usepackage{multirow}
\usepackage{subcaption}
\usepackage{tabularx}
\usepackage{adjustbox}
\usepackage{xcolor}
\usepackage{colortbl}
\usepackage{siunitx}
\usepackage{marvosym}
\usepackage{listings}
\usepackage{tikz}
\usepackage{pgfplots}
\pgfplotsset{compat=1.17}
\usetikzlibrary{arrows.meta,calc,positioning,shapes.geometric,fit,backgrounds,decorations.pathreplacing,patterns}
\usepackage{microtype}
\usepackage{wrapfig}
\usepackage{needspace}
\usepackage[most]{tcolorbox}
\usepackage[scaled=0.92]{inconsolata}
\definecolor{codebg}{HTML}{F2F3F5}
\newtcbox{\inlinecode}{on line, boxsep=0pt, left=2pt, right=2pt, top=1pt, bottom=1pt,
  colback=codebg, colframe=codebg, arc=1.5pt, boxrule=0pt}
\DeclareRobustCommand{\texttt}[1]{\inlinecode{\ttfamily #1}}
\usepackage{url}

\definecolor{grouplab}{HTML}{6E6E73}
\definecolor{oursrow}{HTML}{F2EFF7}
\definecolor{muted}{HTML}{666666}
\newcommand{\ours}{\rowcolor{oursrow}}

\newcommand{\ptabsize}{\fontsize{8}{9}\selectfont}
\newcommand{\gdgroup}[2]{\rowcolor{black!10}[0pt][0pt]\multicolumn{#1}{@{}l@{}}{\textit{#2}}\\}
\newcommand{\gdsecond}[1]{\underline{#1}}
\usepackage{pifont}
\newcommand{\cmark}{\ding{51}}
\newcommand{\pmark}{\ding{55}}

\renewcommand{\arraystretch}{1.15}

\newsavebox{\panelbox}

\tcbset{panel/.style={colback=titleboxbg, colframe=titleboxborder, boxrule=0.5pt, arc=2mm,
  left=6pt, right=6pt, top=5pt, bottom=5pt, before skip=0pt, after skip=0pt,
  valign=center, fontupper=\scriptsize}}

\definecolor{undcolor}{HTML}{5FA8A0}
\definecolor{gencolor}{HTML}{9C7BB5}
\definecolor{jointcolor}{HTML}{D95F59}
\definecolor{barbase}{HTML}{CFCAD9}
\definecolor{barours}{HTML}{6E59B3}
\definecolor{zsok}{HTML}{2F855A}
\definecolor{zsno}{HTML}{B7312C}
\definecolor{ink}{HTML}{18354D}
\definecolor{light}{HTML}{F5F6FA}
\definecolor{cobj}{HTML}{2B6CB0}
\definecolor{cgoal}{HTML}{C05621}
\definecolor{cspa}{HTML}{286B72}
\definecolor{call}{HTML}{5A4888}
\definecolor{codekw}{HTML}{4B3A8C}
\definecolor{codestr}{HTML}{556B2F}
\definecolor{codecom}{HTML}{7A8288}

\newlength{\sbarw}
\newcommand{\dualbar}[4]{%
  \begin{tikzpicture}[baseline={(0,4.4pt)},x=\sbarw,y=1pt]
    \fill[barbase] (0,7.6) rectangle ({#1},13.4);
    \node[anchor=west,inner sep=0pt,xshift=2.5pt,font=\fontsize{7.5}{8}\selectfont,text=muted] at ({#1},10.5) {#2};
    \fill[barours] (0,0) rectangle ({#3},5.8);
    \node[anchor=west,inner sep=0pt,xshift=2.5pt,font=\fontsize{7.5}{8}\selectfont] at ({#3},2.9) {\textbf{\textcolor{barours}{#4}}};
  \end{tikzpicture}}

\definecolor{findingblue}{RGB}{45,82,135}
\definecolor{findingbg}{RGB}{240,246,255}
\newtcolorbox{findingbox}{
    colback=findingbg,
    colframe=findingblue,
    boxrule=0.8pt,
    arc=3mm,
    left=7pt,
    right=7pt,
    top=5pt,
    bottom=5pt,
    boxsep=0pt,
    before skip=6pt,
    after skip=6pt
}
\newcommand{\shows}[1]{%
  \begin{findingbox}
    \textcolor{findingblue}{\textbf{Finding.}} #1
  \end{findingbox}
}

\lstdefinestyle{roampy}{
  language=Python,
  basicstyle=\ttfamily\fontsize{6.0}{7.4}\selectfont,
  keywordstyle=\color{codekw}\bfseries,
  stringstyle=\color{codestr},
  commentstyle=\color{codecom}\itshape,
  emph={build,goal,check,counterexamples},
  emphstyle=\bfseries,
  columns=fullflexible,
  keepspaces=true,
  showstringspaces=false,
  aboveskip=0pt,
  belowskip=0pt,
  frame=none
}

\newcommand{\sysname}{RoMA}
\newcommand{\modelname}{RoMA-27B}
\newcommand{\papersecref}[1]{Section~\ref{#1}}
\newcommand{\paperfigref}[1]{Figure~\ref{#1}}
\newcommand{\papertabref}[1]{Table~\ref{#1}}

\definecolor{titleboxbg}{HTML}{F3F0FA}
\definecolor{titleboxborder}{HTML}{CFC4E8}
\definecolor{titleink}{HTML}{2E2A36}
\definecolor{linkviolet}{HTML}{6E59B3}
\definecolor{urlmagenta}{HTML}{C93D8E}
\colorlet{dahaccent}{linkviolet}
\colorlet{dahaccentfill}{linkviolet}
\hypersetup{
    colorlinks=true,
    linkcolor=linkviolet,
    citecolor=linkviolet,
    urlcolor=urlmagenta,
    pdftitle={RoMA: VLM as Universal Robot Policy via Environment Scaling},
    pdfauthor={Siyuan Hu, Haiwen Diao, Kairui Hu, Weichen Fan, Ziwei Liu}
}
\fancypagestyle{roam}{%
  \fancyhf{}%
  \fancyhead[R]{\sffamily\fontsize{8.5}{10}\selectfont\color{muted}RoMA: VLM as Universal Robot Policy via Environment Scaling}%
  \fancyfoot[C]{\thepage}%
}
\begin{document}
\thispagestyle{roam}

\begin{tcolorbox}[
    colback=titleboxbg,
    colframe=titleboxborder,
    boxrule=0.55pt,
    arc=8mm,
    width=\textwidth,
    left=20pt,
    right=20pt,
    top=18pt,
    bottom=16pt,
    before skip=0pt,
    after skip=14pt
]
\color{titleink}

{\color{black}\sffamily\bfseries\fontsize{19}{23}\selectfont
RoMA: VLM as Universal Robot Policy\\
via Environment Scaling\par}
\vspace{0.80em}

{\color{black}\sffamily\bfseries\fontsize{10}{13}\selectfont\raggedright
Siyuan Hu,\enspace Haiwen Diao,\enspace Kairui Hu,\enspace Weichen Fan,\enspace Ziwei Liu\textsuperscript{\ensuremath{\dagger}}\par}
\vspace{0.35em}

{\fontsize{10.2}{12.4}\selectfont\raggedright
S-Lab, Nanyang Technological University\par}
\vspace{0.2em}
{\fontsize{8.4}{10.2}\selectfont
\textsuperscript{\ensuremath{\dagger}}Corresponding author: ziwei.liu@ntu.edu.sg\par}
\vspace{0.70em}

\begingroup
\fontsize{10.5}{13.2}\selectfont
\setlength{\parindent}{0pt}
\setlength{\parskip}{4pt plus 1pt minus 1pt}
Everyday world knowledge, coding, 2D grounding, and mathematics already give vision-language models (VLMs) the capabilities to model 3D physical scenes in code and infer 3D geometry from RGB images. We introduce \textbf{\sysname{}}, showing that VLMs can turn these native capabilities into both an agentic world model and an end-to-end robot policy. \textbf{(i) VLM as agentic world model:} a VLM coding agent synthesizes 5{,}000 executable simulation environments, each with an automatically verified success check, enabling \modelname{} to train with \textbf{zero external robot data or human demonstrations}. \textbf{(ii) VLM as robot policy:} post-training Qwen3.8-27B inside these environments activates \textbf{closed-loop physical reasoning}, combining native 2D grounding, calibrated math over numerical camera and robot inputs, and iterative feedback to plan primitive actions without external perception tools while preserving multimodal and mathematical reasoning on seven benchmarks. \textbf{(iii) Zero-shot transfer:} \modelname{} transfers directly to \textbf{unseen simulators, hardware setups, and the real world}. Scaling the pool from 100 to 5{,}000 environments lifts LIBERO-Pro success from 11.07\% to 56.07\%, with Spatial reaching 70.8\% and surpassing policies trained directly with RL on LIBERO. On RoboLab-120, \modelname{} achieves the highest overall success at 39.3\% and a 53.0 Score, outperforming both DROID-trained policies and Isaac Sim RL models. In the real world, it achieves 76.7\% success across 30 trials on high-precision tasks including tight-clearance insertion, upright stacking, and multi-stage manipulation. These results show that VLMs can both model physical worlds and act zero-shot via environment scaling, offering a direct path toward universal robot policies.
\endgroup

\vspace{0.5em}
{\sffamily\bfseries\fontsize{8}{11}\selectfont
Project Page:\quad\url{https://vlm-as-policy.github.io/RoMA}\par}
\vspace{0.2em}
{\sffamily\bfseries\fontsize{8}{11}\selectfont
GitHub Repo:\quad\url{https://github.com/vlm-as-policy/RoMA}\par}
\vspace{0.2em}
{\sffamily\bfseries\fontsize{8}{11}\selectfont
Model Weight:\quad\url{https://huggingface.co/collections/h-siyuan/roma}\par}
\vspace{0.2em}

\end{tcolorbox}

\begin{figure}[t]
\centering
\includegraphics[width=\linewidth]{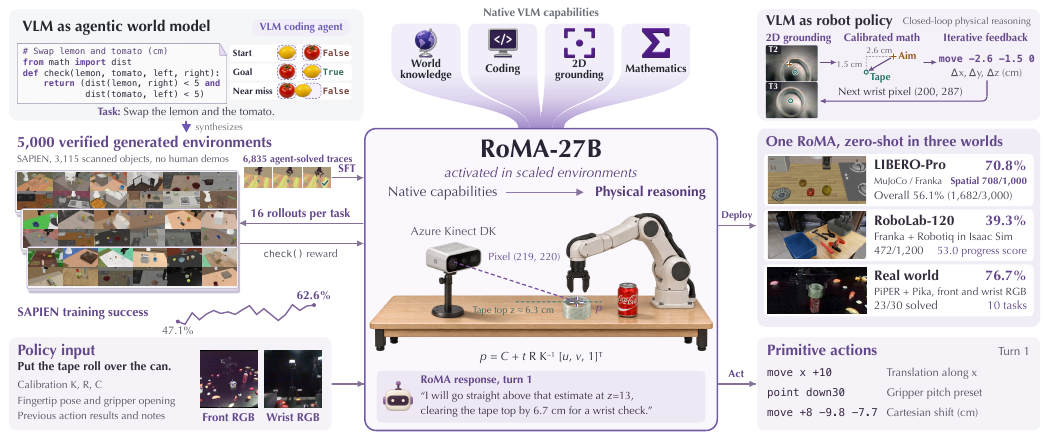}
\caption{\textbf{Overview of \sysname{}.} Starting from a general VLM's native world knowledge, coding, 2D grounding, and mathematics, \sysname{} pairs a VLM coding agent that synthesizes 5{,}000 verifiable SAPIEN environments with a VLM policy post-trained inside them via SFT and GRPO RL using zero external robot data. Activated for closed-loop physical reasoning over calibrated RGB views and primitive Cartesian actions, a single \modelname{} model transfers zero-shot to LIBERO-Pro (MuJoCo), RoboLab-120 (Isaac Sim), and a real PiPER + Pika arm.}
\label{fig:overview}
\end{figure}

\begin{figure}[!t]
\centering
\includegraphics[width=\linewidth]{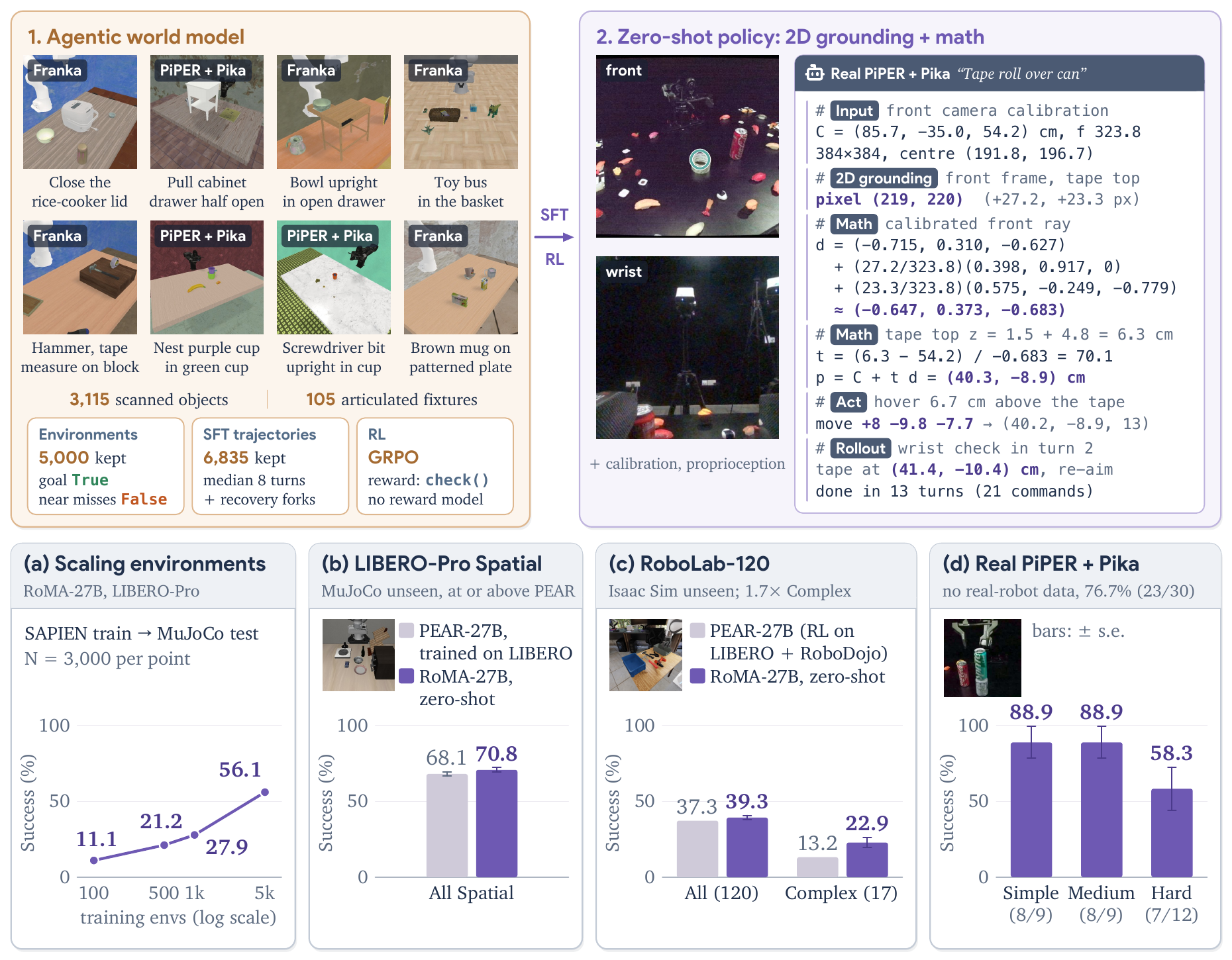}
\caption{\textbf{A VLM synthesizes the worlds \modelname{} trains in, and \modelname{} then acts zero-shot in worlds it has never seen.} (1)~A coding agent synthesizes SAPIEN tasks whose \texttt{check()} graders verify worlds, filter SFT trajectories, and reward GRPO RL. (2)~On turn 1 of ``Tape roll over can'' on a real PiPER + Pika arm, \modelname{} grounds the roll in the front image and intersects its calibrated ray with its top plane without perception tools. (a--d)~Zero-shot transfer across unseen simulators, scenes, and real hardware; error bars show binomial $\pm\text{s.e.}$; PEAR-27B RoboLab-120 follows Table~5 of \citet{pear}, with full splits in Tables~\ref{tab:liberopro}--\ref{tab:real}.}
\label{fig:teaser}
\end{figure}

\section{Introduction}
\label{sec:intro}

Scaling robot learning toward general-purpose manipulation faces two bottlenecks: acquiring diverse, verifiable interaction experience at scale and building policies that transfer across embodiments and environments. In \textbf{\sysname{}} (\paperfigref{fig:overview}), a VLM coding agent resolves the first by synthesizing verifiable simulation environments, and a VLM policy post-trained in them resolves the second by transferring zero-shot across unseen simulators, hardware setups, and the real world.

Synthesizing environments with a coding agent breaks the scale bottleneck of hand-built robot benchmarks. Robot learning depends on executable tasks with programmatic success checks, yet existing suites are written by hand and stop at tens to a few hundred tasks~\citep{libero,robocasa,robodojo}. As an agentic world model, the coding agent writes each task as a self-contained simulation program, inspects rendered views, and revises the code until its success check passes, removing that ceiling so any policy can be rolled out and graded. The same verified success check drives every stage of training. It (i) accepts or rejects the synthesized environment, (ii) filters the trajectories the coding agent solves for supervised fine-tuning, and (iii) supplies the binary reward for reinforcement learning. Because the coding agent constructs every scene, instruction, and check from scratch over 3D asset libraries, the resulting environments share no demonstration, scene file, or initial state with any evaluation benchmark and scale with agent calls and simulators.

Conventional manipulation policies tie their weights to a fixed hardware setup, absorbing specific camera poses and arm kinematics from training data. A VLM policy breaks this coupling through closed-loop physical reasoning grounded in 2D perception and math. At each step, it grounds a target pixel in its visual observations, intersects its calibrated camera ray with a known support plane or a second camera's ray to compute the 3D position, and reasons out the next actions. Post-training in the synthesized environments activates these native grounding and math abilities for closed-loop control through supervised fine-tuning on agent-solved trajectories and reinforcement learning against the verified checks. Because camera calibration and proprioception are given numerically in the prompt, a new hardware setup or physics engine simply updates those values, enabling zero-shot transfer.

This contrasts \sysname{} with two prevailing paradigms (Appendix~\ref{app:comparison}): (i) \textbf{vision-language-action models (VLAs)} rely on teleoperation or hand-crafted suites and absorb camera and arm geometry into policy weights; (ii) \textbf{tool-harness agents} rely on external perception tools whose errors lie outside the VLM's end-to-end visual reasoning; whereas (iii) \textbf{\sysname{}} uses VLM capabilities on both sides, writing verifiable worlds as code and acting via 2D grounding and calibrated math over numerical inputs.

Our contributions are threefold (\paperfigref{fig:teaser}):
\begin{itemize}
  \item \textbf{VLM as agentic world model (\papersecref{sec:world-model}).}
  With zero external robot data or human demonstrations, the coding agent synthesizes 5{,}000 executable SAPIEN~\citep{sapien} environments over 3{,}115 scanned objects~\citep{omni3d,ycb,gso} and 105 articulated fixtures~\citep{artvip} across Franka and PiPER + Pika arms. Each environment undergoes automated verification of goal, initial-state, and failure counterexample checks, lifting zero-shot LIBERO-Pro success by 45.0 percentage points from 100 to 5{,}000 environments (\paperfigref{fig:scaling}).

  \item \textbf{VLM as robot policy (\papersecref{sec:policy}).}
  Post-trained from Qwen3.8-27B~\citep{qwen38} with its vision encoder frozen, \modelname{} performs closed-loop physical reasoning by combining 2D grounding, calibrated math over numerical camera and robot inputs, and iterative feedback to plan primitive Cartesian and gripper actions. Trained with SFT on 6{,}835 agent-solved trajectories and GRPO~\citep{grpo} RL against the verified checks, it preserves general multimodal and mathematical reasoning across seven benchmarks (\papertabref{tab:general}).

  \item \textbf{Zero-shot transfer to unseen worlds (\papersecref{sec:zero-shot}).}
  \modelname{} transfers zero-shot across unseen simulators, grippers, scenes, and tasks at once. It achieves the highest zero-shot result on LIBERO-Pro~\citep{liberopro} in MuJoCo~\citep{mujoco} while leading every method on Spatial, ranks first overall on RoboLab-120~\citep{robolab} in Isaac Sim~\citep{isaacsim} with an unseen Robotiq gripper, and solves high-precision real-world tasks on a PiPER + Pika arm (Tables~\ref{tab:liberopro}--\ref{tab:real}).
\end{itemize}

\begin{figure}[!t]
\centering
\includegraphics[width=\linewidth]{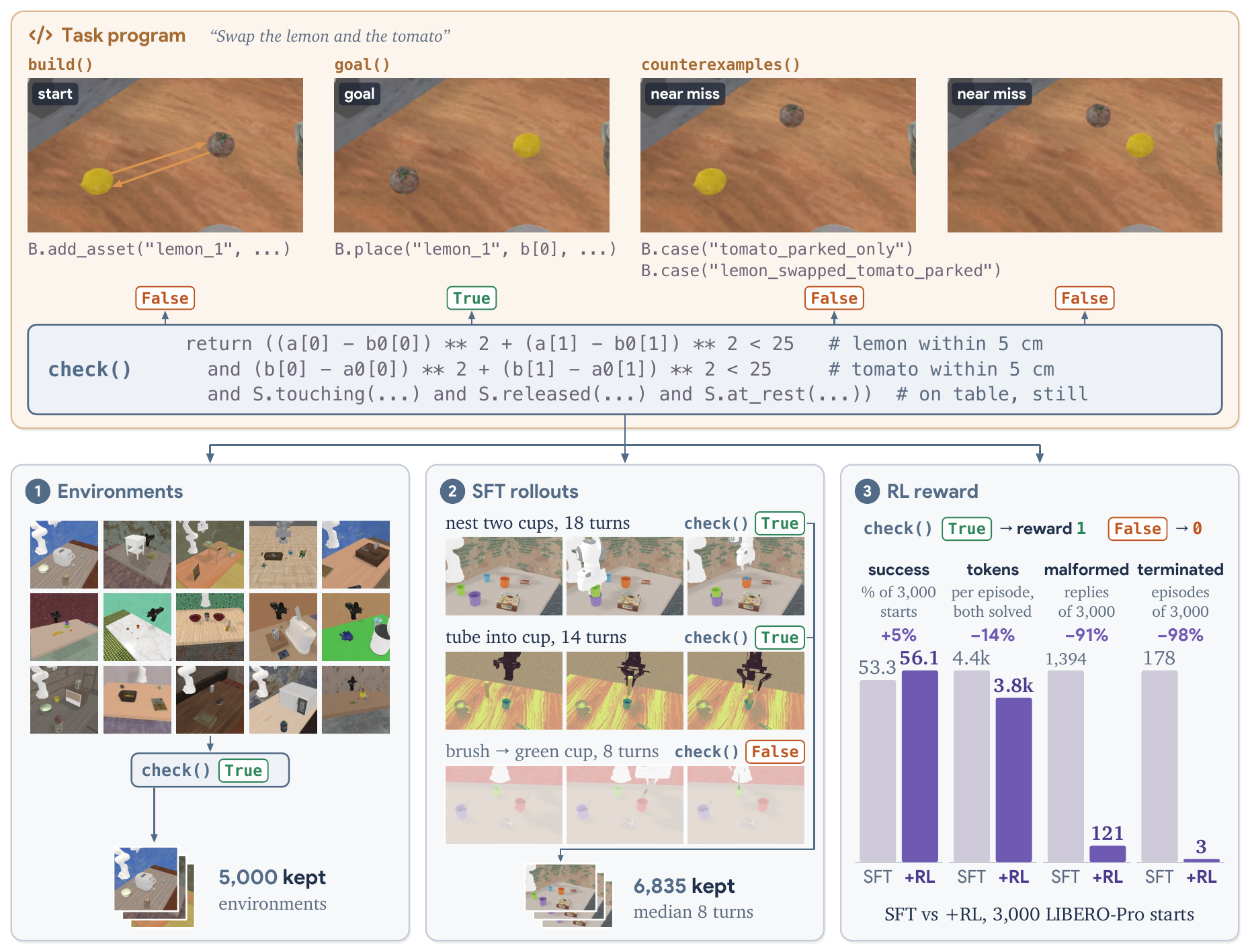}
\caption{\textbf{One generated \texttt{check()} supplies every training signal of \sysname{}.} A synthesized task program (``Swap the lemon and the tomato'') defines \texttt{build()}, \texttt{goal()}, failure \texttt{counterexamples()}, and a physics-grounded \texttt{check()} returning True only on the goal. (1)~These verdicts verify 5{,}000 environments. (2)~Filtering agent rollouts through the policy interface with \texttt{check()} yields 6{,}835 SFT trajectories (median 8 turns). (3)~In GRPO RL, \texttt{check()} supplies the terminal binary reward, lifting LIBERO-Pro successes from 1{,}599 to 1{,}682/3{,}000 while cutting reply tokens by 14.2\% and malformed terminations from 178 to 3.}
\label{fig:method}
\end{figure}

\section{VLM as agentic world model}
\label{sec:world-model}

The agentic world model in \sysname{} uses GPT-6 Astra~\citep{gpt6astra} as a coding agent to synthesize executable SAPIEN environments (\paperfigref{fig:method}). Each program executes any policy's actions under SAPIEN physics and scores the outcome with a programmatic \texttt{check()} function, which (i) verifies the environment, (ii) filters supervised trajectories, and (iii) rewards RL. Below, we describe (i) how the coding agent synthesizes each environment (\papersecref{sec:synthesis}), (ii) how every \texttt{check()} is verified before acceptance (\papersecref{sec:checks}), and (iii) the coverage and scale of the resulting 5{,}000-environment pool (\papersecref{sec:coverage}).

\begin{figure}[t]
\centering
\includegraphics[width=\linewidth]{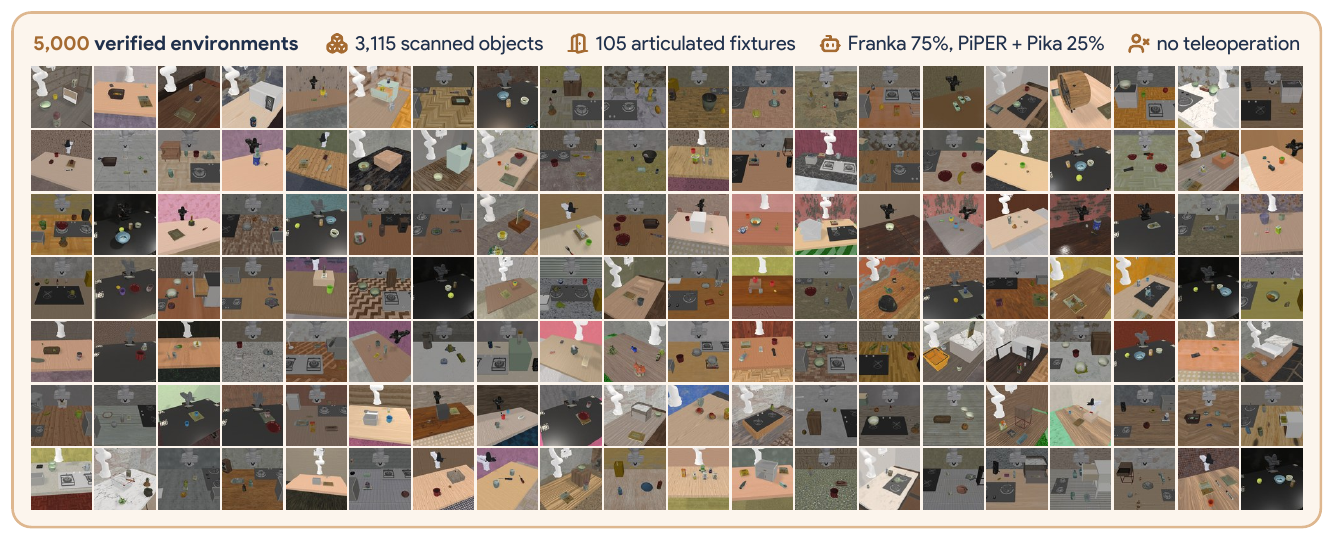}
\caption{\textbf{A coding agent synthesizes 5{,}000 verified environments across two arms.} Each tile is the SAPIEN goal-state render of one verified environment, one distinct instruction per tile, with Franka and PiPER + Pika arms; objects are scans from OmniObject3D (3{,}024), YCB (54), and Google Scanned Objects (37).}
\label{fig:env-pipeline}
\end{figure}

\subsection{Synthesizing an environment}
\label{sec:synthesis}

Representing each environment as one self-contained Python file lets a coding agent build, simulate, and self-verify a manipulation task in one edit-run loop. Written with a scene-builder API and standard arithmetic, the file has five parts: (i) \texttt{build()} builds the start scene, (ii) three paraphrased instructions describe the task unambiguously from the camera views, (iii) \texttt{check()} reads the simulator state and returns a Boolean verdict, (iv) \texttt{goal()} builds the solved state, and (v) \texttt{counterexamples()} builds near misses that test \texttt{check()}. Manipulable objects are 3D scans from OmniObject3D~\citep{omni3d}, YCB~\citep{ycb}, and Google Scanned Objects~\citep{gso}, and articulated cabinets, ovens, microwaves, and trash cans come from ArtVIP~\citep{artvip}; scenes add walls, lighting, textures, and clutter.

An environment tool lets the coding agent catch physical, visual, and kinematic errors before submitting a program, executing each candidate in SAPIEN and returning diagnostic reports. Each report checks (i) whether objects settle without interpenetration, (ii) each object's pixel count in each camera, (iii) whether the arm reaches the relevant poses, and (iv) how \texttt{check()} grades the start state, the goal state, and each counterexample. The agent inspects the rendered start and goal views, revises the program, and submits it for acceptance, taking on average 2.3 measurement reports, 1.3 scene builds, and 1.2 rendered previews per accepted environment. To maintain diversity across the pool, the tool also lists the most similar existing environments with running counts of the relations, objects, and rooms covered so far.

\paragraph{Benchmark separation.} Every training environment is synthesized from scratch in SAPIEN over the asset libraries above, with zero demonstrations, scene files, or initial states from LIBERO or RoboLab. Neither the evaluation simulators (MuJoCo and Isaac Sim) nor any benchmark training split enters \sysname{}'s training, keeping every evaluation in \papersecref{sec:zero-shot} zero-shot.

\subsection{Verifiable success checks}
\label{sec:checks}

A faulty \texttt{check()} corrupts every downstream training signal, poisoning both SFT trajectory filtering and RL rewards. We therefore test every synthesized \texttt{check()} against three conditions before accepting an environment (card~1 of \paperfigref{fig:method}). A verified check returns (i) False on the start state, (ii) True on the constructed goal, and (iii) False on every failure counterexample, such as a partial completion, a one-sided swap, or an unsettled placement. To ground these verdicts in simulator physics, \texttt{check()} reads object positions, displacements from the start, contact pairs, gripper release, and rest velocities directly from SAPIEN. This verification catches frequent bugs during synthesis, with 12.9\% of measurement reports flagging a check that rejected the agent's own goal, accepted a counterexample, or passed the start state. Overall, 20.1\% of environments were revised after such a report. At acceptance, 99.9\% of environments satisfy all three conditions, and 86.1\% include at least one counterexample (median 2).

\subsection{Coverage and scale}
\label{sec:coverage}

Growing the pool from 100 to 5{,}000 environments raises zero-shot LIBERO-Pro success from 11.07\% to 56.07\% (\papersecref{sec:sft-rl}, \paperfigref{fig:scaling}). Across this pool, the synthesized tasks span five relation families: (i) \emph{On top} (40\%), stacking or balancing items on narrow supports; (ii) \emph{Articulated} (26\%), operating doors, lids, knobs, and drawers; (iii) \emph{Inside} (24\%), placing objects into bowls, baskets, pots, and drawers; (iv) \emph{Placement} (6\%), arranging objects around reference items; and (v) \emph{Orientation} (5\%), standing containers upright or aligning their axes. The environments are split across a 7-DoF Franka Panda arm (75\%) and a 6-DoF AgileX PiPER arm with a Pika gripper (25\%), drawing on 3{,}115 scanned objects and 105 articulated fixtures (\paperfigref{fig:env-pipeline}).

\shows{An agentic world model scales verifiable robot training environments with zero external robot data or human demonstrations, lifting zero-shot LIBERO-Pro success from 11.07\% to 56.07\% as the synthesized pool grows from 100 to 5{,}000 environments.}

\section{VLM as robot policy}
\label{sec:policy}

\modelname{} is an end-to-end robot manipulation policy post-trained from Qwen3.8-27B~\citep{qwen38} inside the synthesized environments (\paperfigref{fig:turn}). At every turn, it performs closed-loop physical reasoning by combining 2D visual grounding, calibrated math over numerical camera and robot inputs, and iterative feedback to plan primitive actions. We detail (i) the closed-loop observation and primitive action interface (\papersecref{sec:end-to-end}), (ii) how 2D grounding and calibrated ray arithmetic resolve metric 3D targets (\papersecref{sec:grounding}), (iii) the SFT and RL post-training that activates this reasoning (\papersecref{sec:sft-rl}), and (iv) how post-training preserves general multimodal and mathematical reasoning (\papersecref{sec:general}).

\subsection{End-to-end control with primitives}
\label{sec:end-to-end}

\begin{figure}[!t]
\centering
\includegraphics[width=\linewidth]{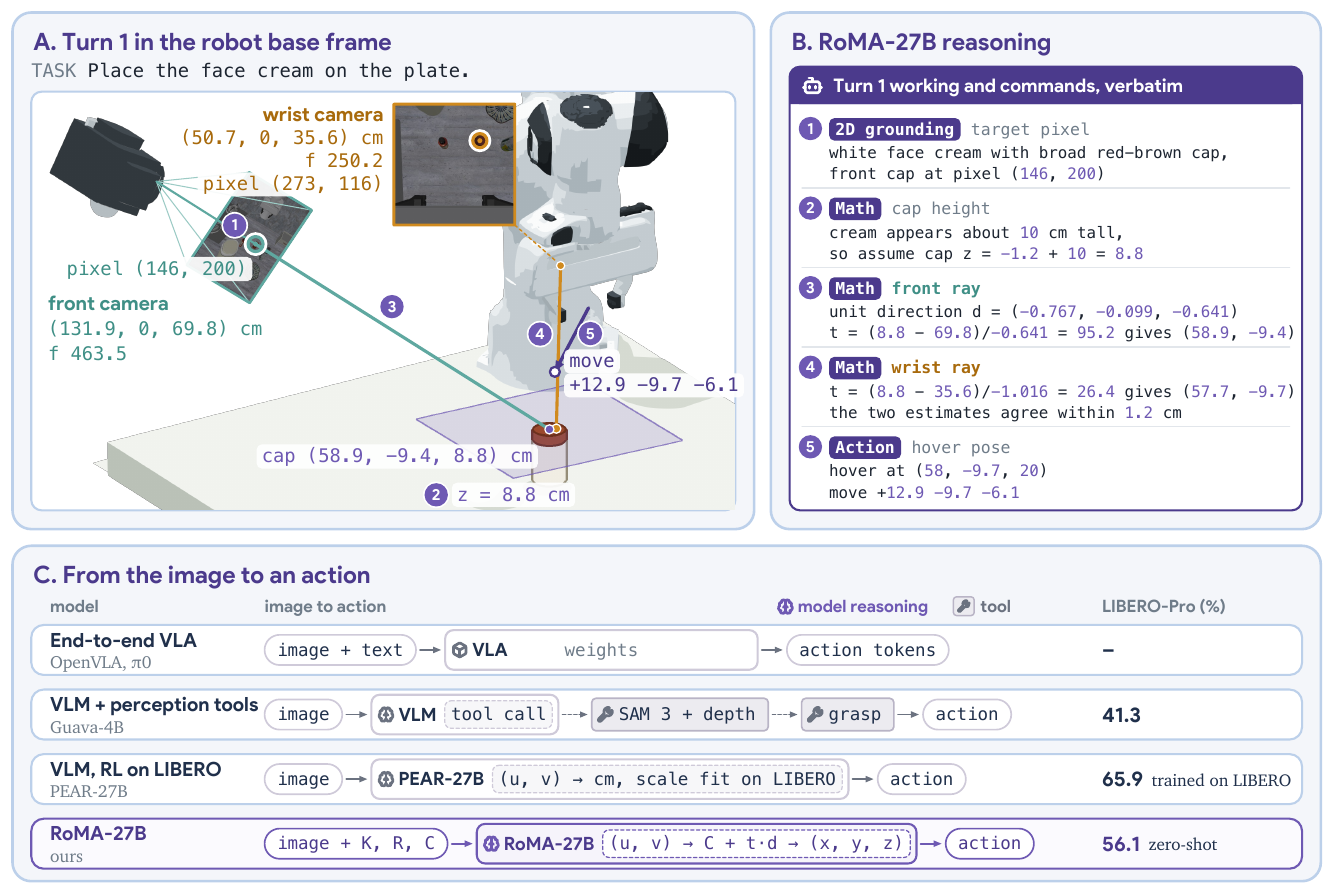}
\caption{\textbf{\sysname{} turns a pixel into a metric action with the 2D grounding and math a VLM already has.} Turn 1 of ``Place the face cream on the plate''. A: the turn drawn in a perspective scene of the robot base frame; the front camera, its frame (drawn as its image plane) and its ray are teal, the wrist camera, its frame and its ray are amber, and what the model computes is violet: (1) the model points to pixel (146, 200); (2) it assumes a cap height of 8.8\,cm; (3) the front ray through that pixel meets this height at the cap, (58.9, $-9.4$, 8.8)\,cm; (4) the wrist ray through pixel (273, 116) of the wrist frame lands within 1.2\,cm of it; (5) the commanded move hovers the gripper 11\,cm above the cap. B: verbatim excerpts of the reply's working and commands, numbered as in A. C: how each model class goes from image to action, with the step that finds the target inside the model's reasoning (dashed) or in tools outside it (grey), and its overall LIBERO-Pro success (\papertabref{tab:liberopro}).}
\label{fig:turn}
\end{figure}

At each turn, \modelname{} processes a structured observation and outputs a single reply containing up to four primitive actions executed directly by a low-level kinematic controller (\paperfigref{fig:turn}). The input bundles (i) the language instruction, (ii) the current front and wrist RGB observations (without frame stacking) alongside each camera's extrinsic pose and focal parameters, (iii) the robot state (fingertip position, approach and finger axes, and gripper aperture), and (iv) feedback from the previous turn, including action outcomes, any contact or kinematic failure cause, a compact turn log, and prior notes. The model's reply has three parts, namely (i) free-form \emph{working} for visual grounding and coordinate calculation, (ii) persistent \emph{notes} carried to the next turn, and (iii) the primitive \emph{actions}. These actions move the fingertip by base-frame Cartesian displacements in centimetres, rotate the wrist, open or close the gripper, or invoke orientation presets, home, wait, and task completion.

The intended action determines what the policy grounds. Grasping a ball targets its centroid, dropping an item into a tape roll targets the inner opening, and standing a tall can upright inside that roll tracks both rim and centre. Acting also verifies each coordinate estimate, because moving to a hover point above an initial front-camera estimate brings the wrist camera overhead and exposes any residual centimetre-level offset before descent. Persistent notes preserve object identities and intermediate progress across multi-stage tasks, while each training turn supervises grounding together with the action that depends on it. Multi-action replies average 2.08 actions per turn across all 1{,}682 solved LIBERO-Pro episodes, cutting VLM calls by 52.0\% over single-action execution (\papertabref{tab:multi-action}).

\begin{table}[t]
\centering
\caption{\textbf{Multi-action replies cut VLM inference calls by over half on LIBERO-Pro.} Mean turns per episode of \modelname{} versus primitive actions executed (i.e., one action per turn) across all 1{,}682 solved episodes of the 3{,}000-rollout LIBERO-Pro evaluation.}
\label{tab:multi-action}
\ptabsize
\begin{tabular}{lrrrr}
\toprule
\textbf{LIBERO-Pro split} & \textbf{Turns / ep.} & \textbf{Actions / ep.} & \textbf{Actions / turn} & \textbf{Inferences saved} \\
\midrule
Spatial (708 solved) & 12.21 & 24.92 & 2.04 & 51.0\% \\
Object (534 solved)  & 10.06 & 21.73 & 2.16 & 53.7\% \\
Goal (440 solved)    & 11.30 & 23.37 & 2.07 & 51.7\% \\
\midrule
\ours \textbf{Overall solved (1{,}682)} & \textbf{11.29} & \textbf{23.50} & \textbf{2.08} & \textbf{52.0\%} \\
\bottomrule
\end{tabular}
\end{table}

\subsection{Grounding and calibrated arithmetic}
\label{sec:grounding}

Pretrained VLMs excel at 2D pixel grounding and elementary arithmetic, yet robot manipulation requires metric 3D coordinates. \sysname{} bridges this gap by reducing 3D perception to (i) grounding 2D pixels and (ii) solving calibrated ray equations. Given a grounded pixel $(u, v)$ in a calibrated camera with intrinsics $\mK$, camera-to-world rotation $\mR_{wc}$, and optical centre $\mathbf{C}_w$, the corresponding 3D point lies along the world-frame ray
\begin{equation}
  \mathbf{p}_w = \mathbf{C}_w + \lambda \, \mR_{wc} \mK^{-1} [u, v, 1]^\top.
  \label{eq:ray}
\end{equation}
A single physical constraint resolves the depth $\lambda$, where (i) an object resting on a tabletop or known surface fixes its height above the support plane, (ii) rays from the front and wrist cameras triangulate 3D coordinates directly, and (iii) the visible gripper fingertips provide an exact metric landmark.

Post-training activates this calculation so that \modelname{} computes 3D object positions to centimetre-level accuracy in an unseen physics engine. We map every 3D coordinate that \modelname{} computes in its first-turn working into the MuJoCo frame and compare it against the ground-truth pose of the nearest object. Across the 2{,}189 LIBERO-Pro rollouts whose first turn computes an explicit object coordinate (typically by intersecting a front-camera ray with an assumed support height), the median horizontal error is 1.5\,cm, dropping to 1.1\,cm on the Object suite with 73\% of all estimates within 3\,cm, and the estimated top height lies a median 1.6\,cm from the true top. Wrist-view refinement closes the remaining gap, bringing the fingertip centre to a median 0.7\,cm from the object's centre when the policy closes the gripper on an Object-suite target, within 2\,cm in 84\% of grasps. \paperfigref{fig:perception} plots the error distribution and contrasts it with PEAR-27B.

\shows{Post-training activates a VLM's native 2D grounding and calibrated math for closed-loop physical reasoning, reaching sub-centimetre grasp accuracy at 0.7\,cm median error after wrist-view refinement in an unseen physics engine, with no LIBERO or MuJoCo training.}

\subsection{Supervised and reinforcement learning at scale}
\label{sec:sft-rl}

To cold-start \modelname{} into closed-loop control without privileged simulator state, we collect supervised trajectories by running the coding agent through the policy's exact interface. It receives the same visual observations, camera intrinsics and extrinsics, proprioceptive state, and primitive action vocabulary, while simulator ground-truth poses are accessed strictly by \texttt{check()} at episode end. Filtering with \texttt{check()} yields 6{,}835 solved trajectories with a median length of 8 turns. We also branch recovery forks from intermediate states of failed or perturbed rollouts and supervise their continuations, teaching the policy to revise earlier coordinate estimates whenever new wrist-camera views or failed actions contradict its notes. SFT runs in bfloat16 precision with sequence length 24{,}576, global batch size 64, peak learning rate $1.4\times10^{-5}$, 5\% linear warm-up, and cosine decay.

\begin{figure}[t]
\centering
\includegraphics[width=\linewidth]{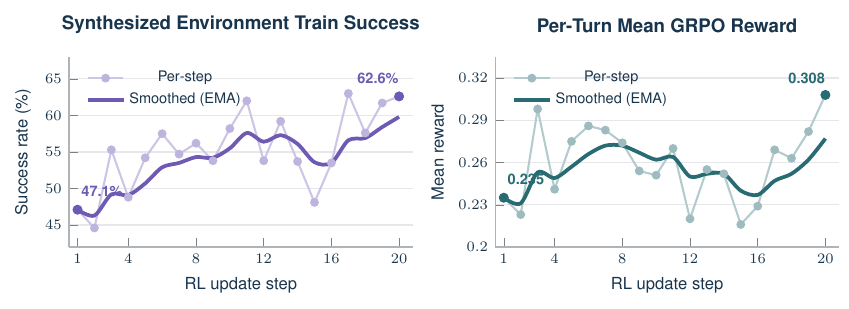}
\caption{\textbf{GRPO steadily improves synthesized-environment training success and per-turn reward.} Left: per-step success on synthesized SAPIEN environments rises from 47.1\% to 62.6\% over 20 RL steps. Right: mean per-turn reward rises from 0.235 to 0.308. Solid curves are exponential moving averages.}
\label{fig:rl-curve}
\end{figure}

Reinforcement learning mirrors the post-training recipe of reasoning language models~\citep{grpo,deepseekr1}, combining (i) verifiable tasks, (ii) filtered reasoning traces, and (iii) group-relative reinforcement. Training runs on 160 GPUs across 20 eight-GPU nodes. At each update step, the SFT-initialized policy samples 16 closed-loop rollouts per task group across 176 groups (oversampling 48 groups beyond the target batch size of 128, for 2{,}816 rollouts per step) from synthesized environments and recovery starts, using its own visual grounding, arithmetic, and primitive actions. Each trajectory receives a binary reward from \texttt{check()} at episode end, and GRPO reinforces the successful rollouts within each group, steadily lifting both synthesized-environment training success (47.1\% to 62.6\%) and per-turn mean GRPO reward (0.235 to 0.308) over 20 update steps (\paperfigref{fig:rl-curve}).

Evaluated across all 3{,}000 LIBERO-Pro rollouts (\papersecref{sec:libero-pro}), RL improves over its SFT initialization along three axes: (i) raising successes from 1{,}599 to 1{,}682 of 3{,}000 (53.30\% to 56.07\%), (ii) cutting reasoning and reply tokens by 14.2\% on the 1{,}268 starts solved by both policies (3{,}806 versus 4{,}438 tokens per episode at the same median of 7 turns), and (iii) reducing malformed replies from 1{,}394 to 121 (91.3\%) and episodes terminated for malformed output from 178 to 3.

\begin{figure}[t]
\centering
\includegraphics[width=\linewidth]{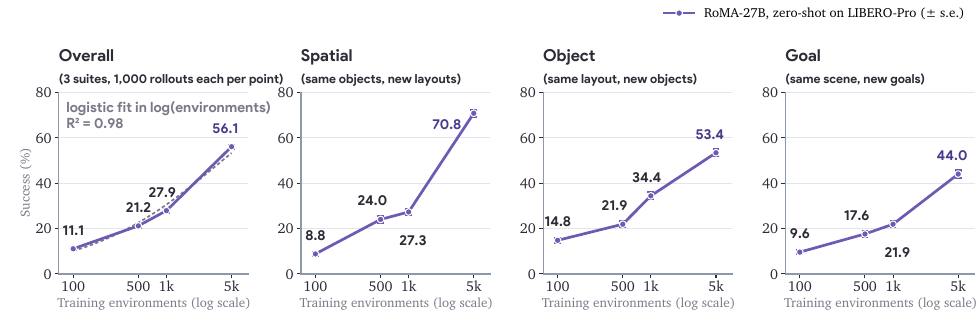}
\caption{\textbf{Scaling training environments from 100 to 5{,}000 raises zero-shot LIBERO-Pro success from 11.07\% to 56.07\%.} Success (\%) for training pools of 100, 500, 1{,}000, and 5{,}000 generated SAPIEN environments (log scale), 3{,}000 rollouts per point and 1{,}000 per suite. Points are measured success with binomial standard errors. Left: Overall, with the logistic fit in the log number of environments ($R^2=0.98$). Right: the Spatial, Object, and Goal suites, each labelled with what it varies, with Spatial rising from 8.8\% to 70.8\%.}
\label{fig:scaling}
\end{figure}

Using the same post-training pipeline and matched training time, growing the synthesized pool from 100 to 5{,}000 drives zero-shot gains across all three LIBERO-Pro suites (\paperfigref{fig:scaling}). Spatial gains the most, rising from 8.8\% to 70.8\%, because its identical objects are distinguished by their spatial relations; Object rises from 14.8\% to 53.4\%, and Goal rises from 9.6\% to 44.0\%.

\subsection{General capabilities}
\label{sec:general}

\begin{table}[t]
\centering
\includegraphics[width=\linewidth]{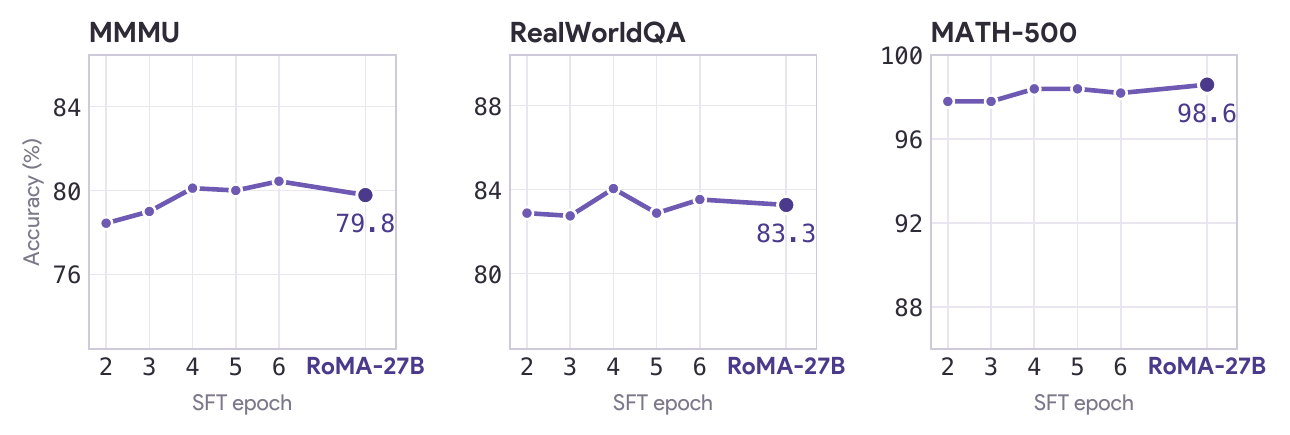}
\captionof{figure}{\textbf{General accuracy holds steady as SFT continues.} Accuracy (\%) on MMMU, RealWorldQA, and MATH-500 after 2 to 6 epochs of SFT and for \modelname{}, from per-checkpoint evaluation runs separate from the paired evaluation in \papertabref{tab:general}. Each panel spans 14 accuracy points.}
\label{fig:general-sft}
\vspace{10pt}
\captionof{table}{\textbf{\modelname{} preserves the general capabilities of its base model under matched paired evaluation.} Accuracy (\%) against Qwen3.8-27B~\citep{qwen38} on the exact same questions using \texttt{evalscope}~1.13.0~\citep{evalscope} and BF16 vLLM~\citep{vllm} thinking-mode sampling ($T=1.0$, top-$p=0.95$, top-$k=20$). Across all seven core multimodal, mathematical, embodied, and instruction-following benchmarks, every paired 95\% bootstrap confidence interval straddles zero; question counts and confidence intervals appear in Appendix~\ref{app:general}.}
\label{tab:general}
\centering
\ptabsize
\setlength{\tabcolsep}{0pt}
\setlength{\sbarw}{6.8cm}
\renewcommand{\arraystretch}{1.85}
\begin{tabular*}{\linewidth}{@{\extracolsep{\fill}}ll@{\hspace{6pt}}l@{\hspace{2pt}}}
\toprule
\textbf{Benchmark} & \textbf{Capability evaluated} & \textcolor{barbase}{\rule{0.7em}{0.7em}}~\textbf{Qwen3.8-27B}\quad \textcolor{barours}{\rule{0.7em}{0.7em}}~\textbf{\modelname{}}\\
\midrule
MATH-500~\citep{math500} & Competition mathematics & \dualbar{0.980}{98.0}{0.986}{98.6}\\
RealWorldQA~\citep{realworldqa} & Real-world visual QA & \dualbar{0.834}{83.4}{0.835}{83.5}\\
MMMU (val)~\citep{mmmu} & College-level multimodal & \dualbar{0.799}{79.9}{0.796}{79.6}\\
MMMU-Pro (standard)~\citep{mmmupro} & Hard 10-option multimodal & \dualbar{0.763}{76.3}{0.754}{75.4}\\
MMMU-Pro (vision) & Vision-only multimodal & \dualbar{0.749}{74.9}{0.739}{73.9}\\
IFBench~\citep{ifbench} & Instruction following & \dualbar{0.733}{73.3}{0.713}{71.3}\\
ERQA~\citep{erqa} & Embodied reasoning & \dualbar{0.650}{65.0}{0.628}{62.8}\\
\bottomrule
\end{tabular*}
\end{table}

Robot post-training can erode a VLM's general reasoning, yet \modelname{} depends on those native abilities to act. Under matched paired evaluation against Qwen3.8-27B (\papertabref{tab:general}), \modelname{} retains its core multimodal, mathematical, embodied, and instruction-following capabilities across seven general benchmarks: 98.6\% on MATH-500 ($+0.6$), 83.5\% on RealWorldQA ($+0.1$), 79.6\% on MMMU ($-0.3$), 75.4\% and 73.9\% on MMMU-Pro standard and vision ($-0.9$ and $-1.0$), 71.3\% on IFBench ($-2.0$), and 62.8\% on ERQA ($-2.3$), with every paired 95\% bootstrap confidence interval straddling zero; question counts and confidence intervals appear in Appendix~\ref{app:general}. Tracking accuracy from 2 to 6 epochs of SFT shows that MMMU, RealWorldQA, and MATH-500 remain steady or improve (\paperfigref{fig:general-sft}). During post-training, we freeze the pretrained vision encoder while updating the vision-language aligner and backbone, and co-train SFT with general multimodal and spatial QA data self-generated by Qwen3.8-27B.

\section{Zero-shot transfer}
\label{sec:zero-shot}

\begin{figure}[t]
\centering
\includegraphics[width=\linewidth]{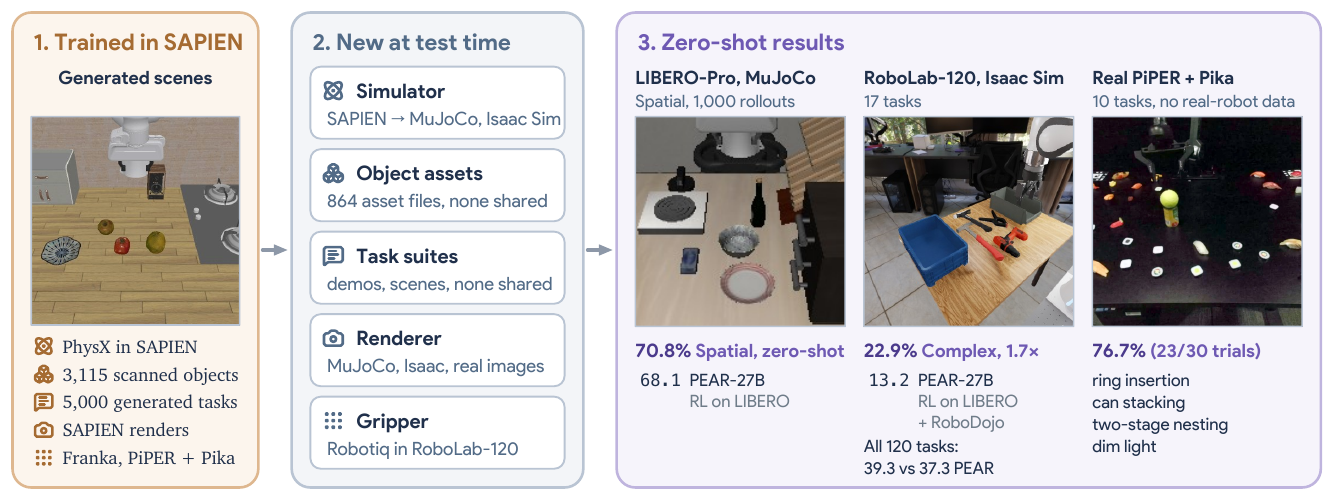}
\caption{\textbf{Trained in SAPIEN, \modelname{} meets a new simulator, a new renderer, new assets, and a new gripper at test time, leading on LIBERO-Pro Spatial and overall on RoboLab-120.} Comparing the synthesized training environments against each unseen evaluation domain highlights every component that changes at test time; none of LIBERO-Pro's 864 asset files, scene files, or initial states appear in training. Across these transfer gaps, \modelname{} reaches 70.8\% on LIBERO-Pro Spatial versus 68.1\% for PEAR-27B trained with RL on LIBERO, 22.9\% versus 13.2\% on the 17 RoboLab-120 Complex tasks (and 39.3\% versus 37.3\% overall) against PEAR-27B trained with RL on LIBERO + RoboDojo, and 23 of 30 real trials on PiPER + Pika with zero real-robot data, including ring insertion, can stacking, and two-stage nesting.}
\label{fig:domain-gap}
\end{figure}

We evaluate \modelname{} zero-shot in test environments that differ from its synthesized SAPIEN training pool along three simultaneous axes (\paperfigref{fig:domain-gap}). Every test environment introduces (i) unseen physics and contact dynamics, spanning MuJoCo~\citep{mujoco} for LIBERO-Pro, Isaac Sim~\citep{isaacsim} for RoboLab-120, and real-world physical contact on hardware, (ii) unseen benchmark tasks, with zero demonstrations, scene files, or initial states from any benchmark entering training, and (iii) unseen observations, with every training frame rendered in SAPIEN and none captured by a real camera. On LIBERO-Pro, every asset is unseen as well: none of its 864 asset files appears in training. Below, we report zero-shot transfer on (i) LIBERO-Pro in MuJoCo (\papersecref{sec:libero-pro}), (ii) RoboLab-120 in Isaac Sim with an unseen Robotiq gripper (\papersecref{sec:robolab}), and (iii) ten real-world tasks on a PiPER + Pika arm (\papersecref{sec:real}).

\subsection{LIBERO-Pro}
\label{sec:libero-pro}

LIBERO-Pro~\citep{liberopro} evaluates manipulation robustness in MuJoCo by perturbing LIBERO tasks. Following the protocol of CaP-X~\citep{capx}, also adopted by Guava~\citep{guava}, we use its object-position and task-instruction perturbations across the Object, Goal, and Spatial suites, totalling 3{,}000 rollouts (500 position and 500 task rollouts per suite; 10 tasks $\times$ 50 initial states per cell). Under LIBERO-Pro's perturbations of objects, initial states, instructions, and environments, VLAs that exceed 90\% on standard LIBERO drop to 0.0\% and continue executing memorized grasps even after the target object is replaced~\citep{liberopro}. Neither MuJoCo, any LIBERO demonstration, scene file, or initial state, nor any of LIBERO-Pro's 864 asset files (verified by hash against 620k training scan files) enters \sysname{}'s training.

\begin{table}[!tb]
\caption{\textbf{Trained without LIBERO or MuJoCo, \modelname{} leads on Spatial and leads all zero-shot models on LIBERO-Pro.} Success rate (\%); parenthetical values in the Object, Goal, and Spatial columns report \emph{position} / \emph{task} perturbation success. Best results per column are in \textbf{bold}. All models are evaluated across the full 3{,}000-rollout protocol of CaP-X~\citep{capx} (500 position and 500 task rollouts per suite), which Guava~\citep{guava} also follows; Guava-4B results are from its Table~3. For fair zero-shot evaluation, nanoAstra-9B receives camera/table-frame alignment with its unused arm masked, and RoboUniverse-35B-A3B receives public Franka kinematics.}
\label{tab:liberopro}
\centering
\footnotesize
\setlength{\tabcolsep}{3pt}
\renewcommand{\arraystretch}{1.15}
\begin{tabular*}{\linewidth}{@{\extracolsep{\fill}}lccccc@{}}
\toprule
Model & Zero-shot & Object {\scriptsize(pos./task)} & Goal {\scriptsize(pos./task)} & Spatial {\scriptsize(pos./task)} & Overall\\
\midrule
\gdgroup{6}{Trained on LIBERO (MuJoCo)}
PEAR-27B~\citep{pear} & \pmark & \textbf{82.1} (88.2/76.0) & \textbf{47.5} (48.2/46.8) & 68.1 (61.0/75.2) & \textbf{65.9}\\
\midrule
\gdgroup{6}{Trained in MuJoCo (no LIBERO tasks)}
Guava-4B~\citep{guava} & \cmark & 60.0 (60/60) & 38.5 (39/38) & 25.5 (25/26) & 41.3\\
\midrule
\gdgroup{6}{Zero-shot (no LIBERO training)}
nanoAstra-9B~\citep{nanoastra} & \cmark & 0.2 (0.2/0.2) & 1.2 (1.8/0.6) & 0.9 (1.6/0.2) & 0.8\\
Qwen3.8-27B (Base)~\citep{qwen38} & \cmark & 2.0 (2.4/1.6) & 2.0 (2.0/2.0) & 5.8 (2.8/8.8) & 3.3\\
RoboUniverse-35B-A3B~\citep{robouniverse} & \cmark & 9.2 (7.6/10.8) & 9.5 (9.2/9.8) & 2.6 (3.8/1.4) & 7.1\\
\modelname{} (Ours) & \cmark & 53.4 (53.6/53.2) & 44.0 (49.8/38.2) & \textbf{70.8} (67.8/73.8) & 56.1\\
\bottomrule
\end{tabular*}
\end{table}

\begin{figure}[t]
\centering
\includegraphics[width=\linewidth]{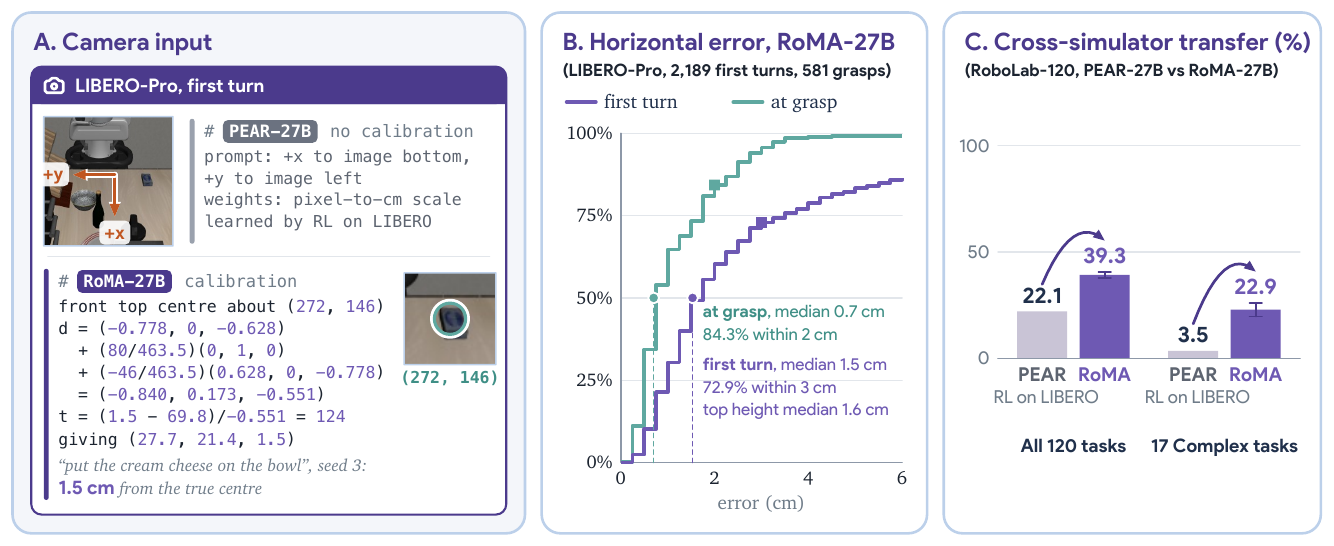}
\caption{\textbf{Calibrated rays carry \modelname{} across simulators; a pixel-to-centimetre mapping learned by RL on LIBERO does not.} Both models see the same LIBERO-Pro front view. PEAR-27B receives the camera only as axis directions in its prompt and learns the mapping of LIBERO's agentview camera through RL on LIBERO, whereas \modelname{} receives numerical camera calibration each turn, grounds a target pixel, and intersects its ray $\mathbf{p} = \mathbf{C} + \lambda\,\mathbf{R}\mathbf{K}^{-1}[u, v, 1]^{\top}$ with the support plane. Over 2{,}189 LIBERO-Pro rollouts, \modelname{}'s first-turn horizontal error has a median of 1.5\,cm, and at 581 Object-suite grasps the gripper sits a median 0.7\,cm from the object. In zero-shot transfer to RoboLab-120 in Isaac Sim with an unseen Robotiq gripper, \modelname{} reaches 39.3\% overall and 22.9\% on the 17 Complex tasks, compared to 22.1\% and 3.5\% for LIBERO-trained PEAR-27B; error bars show binomial standard errors over \modelname{}'s 1{,}200 and 170 episodes.}
\label{fig:perception}
\end{figure}

Trained without LIBERO or MuJoCo, \modelname{} reaches 56.07\% overall across all 3{,}000 LIBERO-Pro rollouts (1{,}682/3{,}000), the highest zero-shot result, and outperforms LIBERO-trained PEAR-27B~\citep{pear} on Spatial overall (70.8\% vs.\ 68.1\%), led by a 6.8-point margin under position perturbations (67.8\% vs.\ 61.0\%) (\papertabref{tab:liberopro}). By contrast, cross-embodiment transfer without calibrated spatial primitives remains challenging for two recent open-weight zero-shot baselines: even with camera-frame alignment and unused-arm masking, nanoAstra-9B~\citep{nanoastra} reaches 0.8\% overall, and RoboUniverse-35B-A3B~\citep{robouniverse} reaches 7.1\% with public Franka kinematics. As illustrated in \paperfigref{fig:perception}, PEAR-27B relies on a pixel-to-centimetre mapping learned for LIBERO's agentview camera; in RoboLab-120, with a new engine, scenes, and gripper, that LIBERO-trained model drops to 22.1\% overall and 3.5\% on Complex tasks, while \modelname{}'s calibrated ray calculation (\papersecref{sec:grounding}) transfers zero-shot at 39.3\% overall and 22.9\% on Complex tasks.

\subsection{RoboLab-120}
\label{sec:robolab}

\begin{table}[!tb]
\caption{\textbf{Never trained in Isaac Sim, \modelname{} leads overall on RoboLab-120.} Task success rate (\%) across all 120 tasks (ten rollouts per task, 1{,}200 episodes for \modelname{}) and stratified by difficulty (64 Simple, 39 Moderate, 17 Complex), alongside RoboLab's graded subtask \emph{Score}. PEAR-27B and leaderboard baselines follow Table~5 of \citet{pear}. Best and second-best results per column are in \textbf{bold} and \underline{underlined}.}
\label{tab:robolab}
\centering
\small
\setlength{\tabcolsep}{3pt}
\renewcommand{\arraystretch}{1.15}
\begin{tabular*}{\linewidth}{@{\extracolsep{\fill}}lccccc@{}}
\toprule
Model & All & Simple & Moderate & Complex & Score\\
\midrule
\gdgroup{6}{Leaderboard policies (trained on DROID)}
Cosmos3-Nano-Policy~\citep{cosmos3} & 36.8 & 40.6 & 35.4 & \textbf{25.3} & \gdsecond{51.9}\\
$\pi_{0.5}$~\citep{pi05} & 28.0 & 29.7 & 31.5 & 13.5 & 43.4\\
DreamZero~\citep{dreamzero} & 25.7 & 26.1 & 30.0 & 14.1 & 39.8\\
GR00T N1.6~\citep{grootn16} & 7.2 & 8.8 & 7.9 & 0.0 & 17.1\\
\midrule
\gdgroup{6}{Trained in Isaac Sim}
PEAR-27B (RL on LIBERO + RoboDojo)~\citep{pear} & \gdsecond{37.3} & \textbf{44.6} & \gdsecond{35.8} & 13.2 & 51.3\\
\midrule
\gdgroup{6}{Zero-shot in Isaac Sim}
PEAR-27B (RL on LIBERO)~\citep{pear} & 22.1 & 28.6 & 19.5 & 3.5 & 35.0\\
\modelname{} (Ours) & \textbf{39.3} & \gdsecond{44.2} & \textbf{38.5} & \gdsecond{22.9} & \textbf{53.0}\\
\bottomrule
\end{tabular*}
\end{table}

RoboLab-120~\citep{robolab} evaluates generalist manipulation in Isaac Sim across 120 tabletop tasks grouped into 64 Simple, 39 Moderate, and 17 Complex tasks, on a Franka arm with a Robotiq parallel-jaw gripper. Isaac Sim, the scenes, the tasks, and the gripper are all unseen in training. We evaluate \modelname{} over 1{,}200 episodes (ten rollouts per task) against (i) PEAR-27B~\citep{pear} both zero-shot (RL on LIBERO) and after adding Isaac Sim RL in RoboDojo~\citep{robodojo}, and (ii) four DROID-trained~\citep{droid2024} leaderboard policies: Cosmos3-Nano-Policy~\citep{cosmos3}, $\pi_{0.5}$~\citep{pi05}, DreamZero~\citep{dreamzero}, and GR00T~N1.6~\citep{grootn16}.

Without any Isaac Sim or Robotiq training data, \modelname{} leads overall in \papertabref{tab:robolab}, reaching 39.3\% across all 120 tasks (472/1{,}200), 44.2\% on Simple, 38.5\% on Moderate, 22.9\% on Complex, and a 53.0 subtask Score. It surpasses all four DROID-trained leaderboard policies, including Cosmos3-Nano-Policy (36.8\%) and $\pi_{0.5}$ (28.0\%), and outperforms PEAR-27B even after it adds Isaac Sim RL in RoboDojo (37.3\%, up from 22.1\% zero-shot). The margin over PEAR-27B widens on long-horizon Complex tasks, where its 22.9\% is $1.7\times$ Isaac-Sim-trained PEAR-27B (13.2\%) and $6.5\times$ zero-shot PEAR-27B (3.5\%).

\shows{General priors and closed-loop physical reasoning enable zero-shot transfer across unseen simulators, scenes, and grippers, leading on LIBERO-Pro Spatial at 70.8\% overall and 67.8\% under position perturbations, and overall on RoboLab-120 at 39.3\% with 22.9\% on Complex.}

\subsection{Real-world evaluation}
\label{sec:real}

\begin{figure}[t]
\centering
\includegraphics[width=\linewidth]{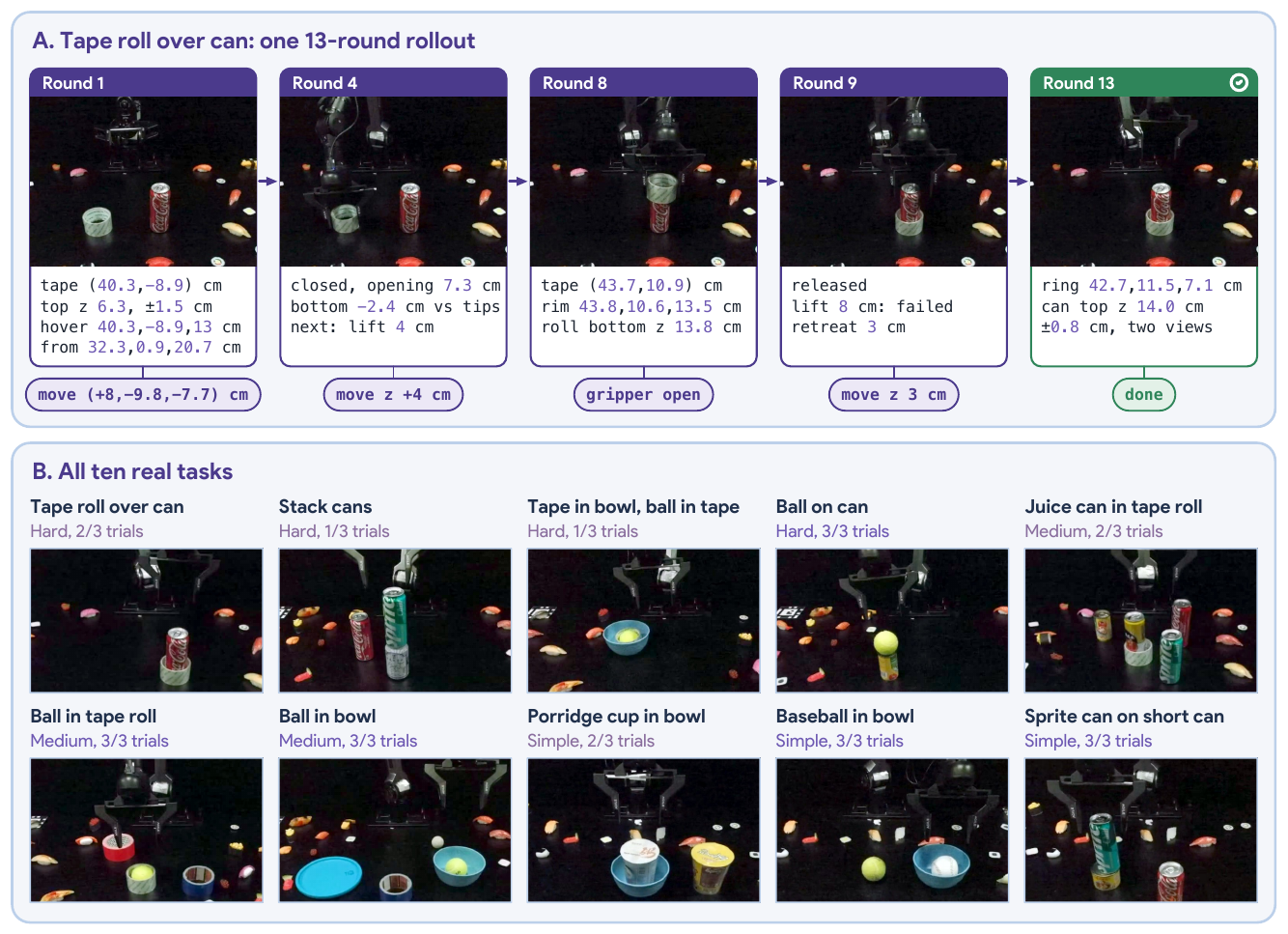}
\caption{\textbf{On a real PiPER + Pika arm with no real-robot training data, \modelname{} succeeds in 23 of 30 trials.} Top: five of the 13 rounds of ``Tape roll over can'' (Hard, 2 of 3 trials succeed), each with a quantity computed that round: the front-ray tape estimate within 1.5\,cm uncertainty, the 7.3\,cm jaw closure, the 0.3\,cm offset between roll and can rim before release, the 13.8\,cm roll-bottom height over the 13.5\,cm lid, the switch from a failed 8\,cm lift to a 3\,cm retreat, and the ring and can top located in two views within 0.8\,cm before reporting done. Bottom: all ten real tasks with successes out of three trials (\papertabref{tab:real}).}
\label{fig:real}
\end{figure}

With zero real-robot observations or trajectories in training, we deploy \modelname{} zero-shot on a real-world AgileX PiPER 6-DoF arm equipped with a Pika gripper, a fixed front RGB camera, and a wrist-mounted RGB camera. Beyond single-object grasping and basic pick-and-place, the ten-task suite tests high-precision placement and multi-stage manipulation, including (i) balancing a tennis ball atop a narrow cylindrical can, (ii) lowering a transparent tape roll concentrically over an upright can with millimetre-scale clearance, (iii) stacking tall cans upright, and (iv) two-stage compositional nesting (\paperfigref{fig:real}); we additionally evaluate a dim-lighting rollout in \papersecref{sec:quality}.

\begin{table}[!tb]
\caption{\textbf{With zero real-robot training data, \modelname{} succeeds in 23 of 30 trials across ten tasks on the PiPER + Pika arm.} Three trials per task, with a fixed front RGB camera and a wrist RGB camera.}
\label{tab:real}
\centering
\footnotesize
\setlength{\tabcolsep}{3pt}
\renewcommand{\arraystretch}{1.15}
\begin{tabularx}{\linewidth}{@{}l>{\raggedright\arraybackslash}Xr@{}}
\toprule
Task & Instruction summary & Success\\
\midrule
Total & & 23 / 30 (76.7\%)\\
\midrule
\gdgroup{3}{Simple tasks}
Baseball in bowl & Put the baseball into the bowl & 3 / 3\\
Sprite can on short can & Put the Sprite can on the short can & 3 / 3\\
Porridge cup in bowl & Put the instant porridge into the bowl & 2 / 3\\
Simple subtotal & & 8 / 9 (88.9\%)\\
\midrule
\gdgroup{3}{Medium tasks}
Ball in tape roll & Place tennis ball inside flat clear tape roll & 3 / 3\\
Ball in bowl & Place tennis ball inside blue bowl & 3 / 3\\
Juice can in tape roll & Stand apple juice can upright inside clear tape roll & 2 / 3\\
Medium subtotal & & 8 / 9 (88.9\%)\\
\midrule
\gdgroup{3}{Hard tasks}
Ball on can & Balance tennis ball on top of cylindrical can & 3 / 3\\
Tape roll over can & Lower clear tape roll over upright Coca-Cola can & 2 / 3\\
Stack cans & Stack Sprite can upright on top of white can & 1 / 3\\
Tape in bowl, ball in tape & Tape roll into blue bowl, then tennis ball into it & 1 / 3\\
Hard subtotal & & 7 / 12 (58.3\%)\\
\bottomrule
\end{tabularx}
\end{table}

Across three trials per task, \modelname{} succeeds in 23 of 30 trials (76.7\%) (\papertabref{tab:real}), reaching (i) 8/9 (88.9\%) on Simple tasks, (ii) 8/9 (88.9\%) on Medium tasks, and (iii) 7/12 (58.3\%) on Hard tasks involving millimetre-scale ring clearance, tall can stacking, and two-stage placement where the first placed object becomes the receptacle for the second. In the 13-round rollout of \paperfigref{fig:real}, this closed-loop physical reasoning governs fine physical details. The policy (i) compensates horizontal drift during a $30^\circ$ pitched descent ($\Delta x = \Delta z \tan 30^\circ$), (ii) accounts for the 2.4\,cm overhang of the grasped tape roll below the fingertips to clear the 13.5\,cm can lid during transport, and (iii) withholds completion until retreating the arm to verify concentric settlement from clear front and wrist views.

\subsection{Analysis}
\label{sec:quality}

We next evaluate inference latency under matched hardware and robustness under dim lighting.

\begin{table}[!tb]
\caption{\textbf{On identical $2\times$H100 (TP=2, BS=1) hardware across all 60 unique LIBERO-Pro tasks, \modelname{} runs $8.9\times$ faster per turn and $3.0\times$ faster end-to-end on solved episodes than PEAR-27B.} Evaluated across all 60 unique tasks (6 cells $\times$ 10 tasks, seed~1). \emph{Turn LLM} reports mean / P90 pure LLM latency per call; \emph{In / Out tok} reports mean prompt and generated tokens per call; \emph{Ep.\ LLM / E2E} reports mean pure LLM and end-to-end episode time; \emph{Solved E2E} reports mean end-to-end time on solved episodes (---: none solved).}
\label{tab:latency}
\centering
\footnotesize
\setlength{\tabcolsep}{3pt}
\renewcommand{\arraystretch}{1.15}
\begin{tabular*}{\linewidth}{@{\extracolsep{\fill}}lccccc@{}}
\toprule
Model & Turn LLM (s) {\scriptsize(mean/P90)} & In / Out tok & Calls/ep. & Ep.\ LLM / E2E (s) & Solved E2E (s)\\
\midrule
PEAR-27B~\citep{pear} & 42.23 / 107.75 & 34.8k / 3{,}258 & 13.5 & 568.1 / 584.3 & 177.6\\
Qwen3.8-27B (Base)~\citep{qwen38} & 40.09 / 49.50 & 3.6k / 3{,}339 & 9.6 & 383.5 / 392.7 & 425.0\\
RoboUniverse-35B-A3B~\citep{robouniverse} & 1.82 / 2.72 & 60.2k / 163 & 110.0 & 200.5 / 325.6 & 189.7\\
nanoAstra-9B~\citep{nanoastra} & \textbf{0.95} / \textbf{1.12} & 2.3k / 178 & 45.0 & \textbf{42.9} / \textbf{118.6} & ---\\
\midrule
\modelname{} (Ours) & 4.75 / 7.00 & 3.8k / 378 & 23.1 & 109.9 / 134.2 & \textbf{58.8}\\
\bottomrule
\end{tabular*}
\end{table}

\paragraph{Inference latency.} On identical $2\times$H100 hardware (TP=2, single-stream batch size 1) across all 60 unique LIBERO-Pro tasks, \modelname{} runs $8.9\times$ faster per turn (4.75\,s vs.\ 42.23\,s mean; 7.00\,s vs.\ 107.75\,s P90, $15.4\times$ faster) and $3.0\times$ faster end-to-end on solved episodes (58.8\,s vs.\ 177.6\,s; 47.2\,s vs.\ 165.4\,s pure LLM) than PEAR-27B at nearly identical effective output rates (79.5 vs.\ 77.1\,tok/s) (\papertabref{tab:latency}), cutting full-set pure LLM time by $5.2\times$ (109.9\,s vs.\ 568.1\,s) and full-set end-to-end time by $4.4\times$ (134.2\,s vs.\ 584.3\,s). This speedup comes from the interface and post-training rather than hardware throughput: single-turn observations with persistent notes keep prompts compact at 3.8k versus 34.8k input tokens, while post-training compresses reasoning into concise calibrated arithmetic at 378 output tokens per turn versus 3{,}258 for PEAR-27B and 3{,}339 for Qwen3.8-27B Base, yielding an $8.4\times$ per-turn speedup over Base. Although RoboUniverse-35B-A3B has shorter individual calls at 1.82\,s, its 110.0 calls and 125.1\,s of simulator and tool overhead per episode make it $2.4\times$ slower end-to-end overall (325.6\,s vs.\ 134.2\,s) and $3.2\times$ slower on solved episodes (189.7\,s vs.\ 58.8\,s). The smaller nanoAstra-9B achieves the lowest latency at 0.95\,s per turn and 118.6\,s end-to-end, but solves 0 of 60 tasks here and reaches only 0.8\% overall on LIBERO-Pro (\papertabref{tab:liberopro}).

\begin{figure}[!tb]
\centering
\includegraphics[width=\linewidth]{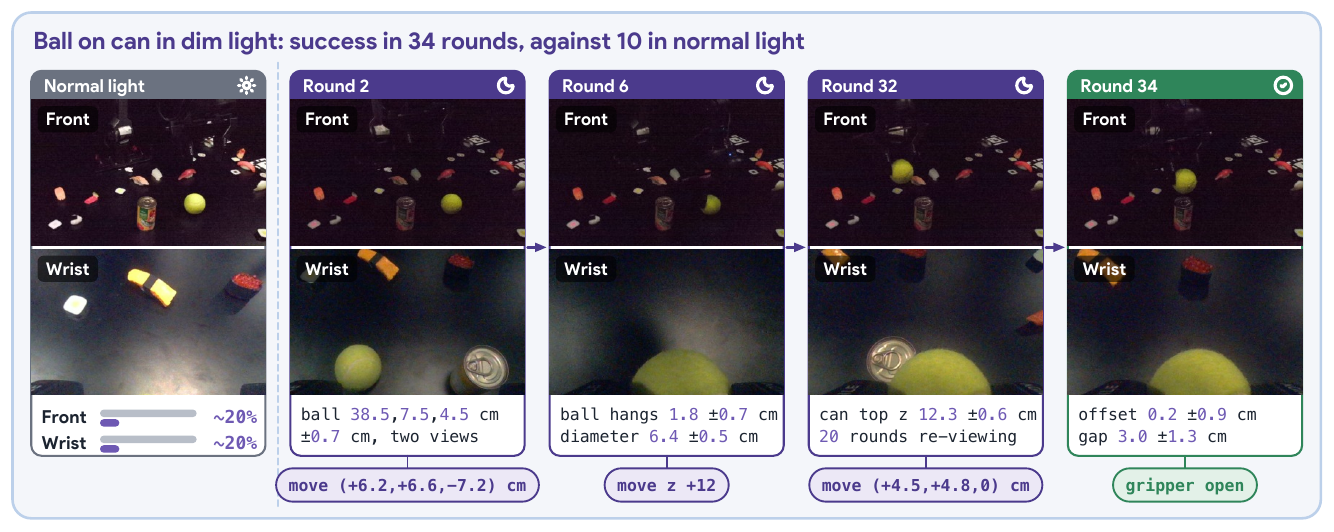}
\caption{\textbf{In dim light, \modelname{} still localizes the ball and the can lid from two views, and spends its extra rounds re-viewing the lid before release.} Left: the same model and task with lights on (10 rounds); bars show that dim pixel intensities of the front and wrist inputs fall to about 20\% of those with lights on. Right: four of the 34 dim-episode rounds with front and wrist inputs, estimates, and commands: two-view ball centre (round 2), ball hang below the fingertips after the grasp (round 6), lid height re-measured after 20 re-viewing rounds (round 32), and axis offset and gap at release (round 34).}
\label{fig:dim}
\end{figure}

\paragraph{Dim lighting.} Even when dim lighting cuts pixel intensities in the front and wrist images to about 20\% of the bright-light frames, \modelname{} still completes Ball on can, a Hard task, in 34 rounds compared to 10 with the lights on, using the same instruction (``put the tennis ball on top of the can.'') and cameras (\paperfigref{fig:dim}). Localization holds in the dark. In round 2 the front and wrist rays through the ball meet at (38.5, 7.5, 4.5)\,cm within 0.7\,cm, and those through the lid at (39.5, $-8.5$, 11.5)\,cm within 0.8\,cm; the jaws close on the 6.4\,cm ball in round 5, and it hangs 1.8\,cm below the fingertips. The extra rounds go to verification. Over the can, the held ball hides the lid from the wrist camera, and in rounds 12 to 31 the policy shifts sideways, backs away, lifts, and changes pitch to see the lid again. In round 32 a backward move exposes the full lid, which both views place at 12.3\,cm within 0.6\,cm. The policy then aligns the ball to within 0.2\,cm of the can axis, descends to a 3.0\,cm gap, and releases.

\section{Discussion and limitations}
\label{sec:discussion}

The results in \papersecref{sec:zero-shot} show that VLMs can both synthesize verifiable training worlds and act zero-shot across unseen simulators and real hardware. Below, we examine the main boundaries of the current system and the open problems that follow.

\paragraph{Toward real-time and dynamic control.}
On identical 2$\times$H100 hardware, \modelname{} cuts per-turn inference latency to 4.75\,s, an 8.9$\times$ speedup over PEAR-27B (\papertabref{tab:latency}), which is sufficient for quasi-static precision tasks such as tight-clearance insertion, upright stacking, and multi-stage manipulation. Reaching 10--50\,Hz reactive control for dynamic tasks, including moving targets, sliding, and compliant force control, remains an open gap. Because per-turn latency is dominated by autoregressive reasoning decode, two complementary paths stand out: (i) shorter reasoning chains and distillation into smaller backbones, and (ii) dual-rate asynchronous execution, where the VLM plans metric geometric targets and verifies physical state while a lightweight high-frequency local controller tracks continuous trajectories and contact forces in real time.

\paragraph{Where zero-shot physical reasoning still fails.}
The benchmark breakdown clarifies what 2D grounding and calibrated ray arithmetic solve and where failures remain. On LIBERO-Pro, \modelname{} leads every method on Spatial (70.8\%), while Object (53.4\%) and Goal (44.0\%) still trail in-domain LIBERO RL (\papertabref{tab:liberopro}); on RoboLab-120 and real hardware, overall success reaches 39.3\% and 23/30, with real-world failures concentrated in Hard tasks (7/12; Tables~\ref{tab:robolab} and~\ref{tab:real}). Calibrated ray math directly resolves metric spatial localization, achieving 1.1\,cm median turn-1 horizontal error and 0.7\,cm median fingertip-to-object error at grasp on the Object suite (\paperfigref{fig:perception}). Remaining failures fall into three narrower regimes: (i) fine-grained contact and grasp geometry under unseen gripper dynamics or tight tolerances, (ii) long-horizon subgoal tracking and active re-observation under occlusion, and (iii) subtle visual-semantic object disambiguation in clutter.

\paragraph{The next scaling frontier of agentic world models.}
Scaling the synthesized pool from 100 to 5{,}000 environments lifts zero-shot LIBERO-Pro success by 45.0 percentage points (\paperfigref{fig:scaling}), showing that task diversity remains a primary driver in this regime. Pushing further will shift the bottleneck from task count to physical richness: extending code-synthesized environments to deformable objects, bimanual and dexterous contact, and mobile manipulation. Equally important is strengthening programmatic verification. In the current pipeline, \texttt{check()} is validated against constructed goal, initial, and counterexample states (with 86.1\% of environments containing at least one counterexample; \papersecref{sec:checks}), which catches major grader bugs but is not a formal completeness proof; richer adversarial testing will matter as tasks grow longer and more contact-rich.

\paragraph{General intelligence and evaluation scope.}
Post-training \modelname{} preserves core reasoning at 98.6\% on MATH-500, 79.6\% on MMMU, and 83.5\% on RealWorldQA (\papertabref{tab:general}), indicating that when manipulation is expressed in a VLM's native visual and mathematical reasoning rather than embodiment-specific motor tokens, embodied control can scale alongside general intelligence rather than competing with it. At the same time, our real-world evaluation currently covers one hardware setup (PiPER + Pika, 10 tabletop tasks, 30 trials) with calibrated cameras. Extending zero-shot physical reasoning to uncalibrated cameras, bimanual or mobile embodiments, and unstructured outdoor environments remains important future work.

\section{Related work}
\label{sec:related}

\paragraph{World models and environment synthesis.}
World models are traditionally learned from experience to predict how observations evolve under actions~\citep{worldmodels,dreamerv3}, a paradigm scaled to interactive video by UniSim~\citep{unisim}, Genie~\citep{genie}, and Cosmos~\citep{cosmoswfm,cosmos3}, and coupled with action prediction in DreamZero~\citep{dreamzero}. A parallel line represents world models as executable code, where language models write and refine environment transitions~\citep{worldcoder,codeworldmodels}. In robot manipulation, benchmarks such as LIBERO~\citep{libero}, RoboCasa~\citep{robocasa}, RoboLab~\citep{robolab}, and RoboDojo~\citep{robodojo} curate tasks and success conditions by hand or with LLM assistance on platforms like robosuite~\citep{robosuite} and Isaac Sim~\citep{isaacsim}. Language models first automated individual components, including task code~\citep{gensim}, proposals and rewards~\citep{robogen,eureka,text2reward}, and 3D layouts~\citep{holodeck}. More recently, coding agents have begun building complete environments, with Gym-Anything for Robotics~\citep{gymanything} constructing 200 audited RL environments across 25 simulations, EmbodiedSWE~\citep{embodiedswe} turning coding-agent solutions into VLA training data, and Real2Gym~\citep{real2gym}, Video2Policy~\citep{video2policy}, and RoboUniverse-10K~\citep{robouniverse} reconstructing tasks from videos. \sysname{} extends this direction by using a coding agent as an agentic world model to synthesize 5{,}000 verified SAPIEN environments with zero external robot data, enabling zero-shot transfer to unseen simulators.

\paragraph{VLM grounding and robot-use agents.}
Pretrained VLMs such as Molmo~\citep{molmo} ground language to 2D pixels. Robotics works use it by fine-tuning for affordance points~\citep{robopoint}, selecting visual marks overlaid on images~\citep{pivot,moka}, or training on synthetic metric spatial QA~\citep{spatialvlm}. Another line composes perception and motion APIs through code or tool harnesses, from Code as Policies~\citep{cap}, VoxPoser~\citep{voxposer}, and CaP-X~\citep{capx} to recent harnesses using native robot software~\citep{openrua}, semantic action units~\citep{showharness}, unified open-model interfaces~\citep{ura}, stateful code~\citep{etm}, and tool-calling distillation~\citep{guava}. \modelname{} instead integrates grounded pixels, calibrated ray arithmetic, and iterative feedback into closed-loop physical reasoning, outputting primitive actions without external perception or grasp tools.

\paragraph{Post-training and zero-shot transfer for manipulation.}
Generalist policies broaden coverage by randomizing visual appearance and dynamics~\citep{domainrand,dynrand,rubikscube} or by training vision-language-action (VLA) models on large multi-embodiment demonstration pools~\citep{rt2,openvla,pi0,pi05,openx,octo}. Recent work also post-trains VLAs and open-weight VLMs with reinforcement learning, including in-simulator RL for VLAs~\citep{rl4vla,simplevlarl}, harness-based SFT and RL~\citep{captain,agenticroborl,roboharnessrl}, pure RL on LIBERO and RoboDojo~\citep{pear}, self-bootstrapped primitive control~\citep{zap}, coding-agent distillation~\citep{robouniverse}, and large-scale real-robot fine-tuning mixed with VQA data~\citep{nanoastra}. \sysname{} post-trains \modelname{} with SFT and RL inside its agent-synthesized SAPIEN environments using zero external robot data or benchmark-specific assets, and transfers zero-shot across unseen simulators, hardware setups, and the real world.

\section{Conclusion}
\label{sec:conclusion}

We introduced \sysname{}, demonstrating that everyday world knowledge, coding, 2D grounding, and mathematics already allow VLMs to serve both as an agentic world model and as an end-to-end robot policy. On the modeling side, a VLM coding agent synthesizes 5{,}000 executable SAPIEN environments with automatically verified success checks, scaling verifiable physical experience with zero external robot data or human demonstrations. On the acting side, post-training Qwen3.8-27B inside those environments activates closed-loop physical reasoning over 2D visual observations and numerical camera and robot inputs, transferring zero-shot across unseen simulators, grippers, and a real robot arm while preserving general multimodal and mathematical reasoning. Extending agentic world models to richer contact physics and closing the gap toward real-time closed-loop control offer a direct path for VLMs to both model the world and act in the wild.

\setlength{\bibsep}{2pt plus 1pt}
\bibliography{references}
\bibliographystyle{plainnat}

\clearpage
\appendix

\section{Comparison with concurrent VLM robot policies}
\label{app:comparison}

\papertabref{tab:comparison} summarizes the training data, environment construction, perception interface, post-training objective, and cross-simulator evaluation scope of \sysname{} alongside five concurrent VLM-based robot manipulation frameworks.

\begin{table}[!htb]
\caption{\textbf{Comparison of \sysname{} against recent VLM robot policies and agentic robotics frameworks.} \sysname{} is the only approach combining zero real-robot data, agent-synthesized environments at scale, end-to-end visual reasoning without external perception tools, multi-turn online RL, and zero-shot cross-simulator transfer. $^{*}$PEAR's full model trains with RL in both MuJoCo (LIBERO) and Isaac Sim (RoboDojo); its LIBERO-only ablation transfers zero-shot to Isaac Sim at 22.1\% (\papertabref{tab:robolab}).}
\label{tab:comparison}
\centering
\footnotesize
\setlength{\tabcolsep}{4pt}
\renewcommand{\arraystretch}{1.15}
\begin{tabularx}{\linewidth}{@{}Xccccc@{}}
\toprule
Method & Zero Real Data & Synthesized Envs & No Percep.\ Tools & Online RL & Cross-Sim Zero-Shot\\
\midrule
PEAR~\citep{pear} & \cmark & \pmark & \cmark & \cmark & \pmark$^{*}$\\
RoboUniverse~\citep{robouniverse} & \pmark & \cmark & \pmark & \pmark & \cmark\\
nanoAstra~\citep{nanoastra} & \pmark & \pmark & \cmark & \pmark & \cmark\\
RoboHarness-RL~\citep{roboharnessrl} & \cmark & \pmark & \pmark & \cmark & \pmark\\
ZAP~\citep{zap} & \cmark & \pmark & \pmark & \cmark & \pmark\\
\midrule
\ours \textbf{\sysname{} (Ours)} & \cmark & \cmark & \cmark & \cmark & \cmark\\
\bottomrule
\end{tabularx}
\end{table}

\section{Policy interface, prompt structure, and primitive action space}
\label{app:interface}

At every turn, \modelname{} receives a single-turn structured observation without frame stacking and outputs a structured reply containing internal reasoning, persistent notes, and up to four primitive commands. Execution proceeds sequentially through the listed commands; if any command fails due to inverse kinematics or contact limits, execution stops immediately, remaining commands in the turn are marked \texttt{SKIPPED}, and the exact failure reason is returned in the next turn's observation.

\paragraph{Structured observation and reply format.}
Each prompt supplies four fields: (i) the natural-language task instruction; (ii) the current front and wrist RGB frames ($384\times 384$) together with each camera's focal length, principal point, optical centre $\mathbf{C}_w$, and camera-to-world rotation $\mathbf{R}_{wc}$ in the robot base frame; (iii) proprioceptive state, including fingertip position, approach and finger axes, and gripper aperture; and (iv) previous-turn feedback, consisting of per-command execution outcomes, any failure diagnostics, a compact turn log, and the previous turn's \texttt{NOTES}. The policy replies in three delimited blocks: (i) \texttt{<think>...</think>} working for 2D pixel grounding, calibrated ray arithmetic, and clearance checks (stripped before passing history to subsequent turns); (ii) \texttt{NOTES:} carrying concise object coordinates and subgoal progress to the next turn; and (iii) \texttt{COMMANDS:} listing 1 to 4 primitive actions from \papertabref{tab:actions}.

\begin{table}[!htb]
\caption{\textbf{Primitive action vocabulary of \modelname{}.} All displacements are expressed in centimetres in the robot base frame and all wrist rotations in degrees. Up to four primitives may be issued in a single turn.}
\label{tab:actions}
\centering
\footnotesize
\setlength{\tabcolsep}{4pt}
\renewcommand{\arraystretch}{1.15}
\begin{tabularx}{\linewidth}{@{}l>{\raggedright\arraybackslash}Xl@{}}
\toprule
Command & Semantics & Unit / Arguments\\
\midrule
\texttt{move dx dy dz} & Translate fingertip by $(\Delta x, \Delta y, \Delta z)$ in the robot base frame & cm (3 floats)\\
\texttt{move <axis> $\pm$d} & Translate fingertip along a single base axis (\texttt{x}, \texttt{y}, or \texttt{z}) & axis, cm\\
\texttt{rotate <axis> $\pm\theta$} & Rotate wrist around \texttt{yaw}, \texttt{pitch}, or \texttt{roll} & axis, degrees\\
\texttt{point <preset>} & Set gripper pitch to a canonical preset (\texttt{down}, \texttt{down15}, \texttt{down30}, \texttt{down45}) & preset name\\
\texttt{gripper open / close} & Open or close the parallel-jaw gripper & binary state\\
\texttt{home} & Move arm to its neutral home configuration & none\\
\texttt{wait} & Hold current pose and re-observe to verify contact or settling & none\\
\texttt{done} & Declare task completion and terminate the episode & none\\
\bottomrule
\end{tabularx}
\end{table}

\section{Post-training hyperparameters and scaling protocol}
\label{app:training}

\papertabref{tab:hparams} summarizes the supervised fine-tuning (SFT) and group-relative reinforcement learning (GRPO RL) configurations used to post-train \modelname{} from Qwen3.8-27B~\citep{qwen38}. In both stages, the pretrained vision encoder remains frozen while the vision-language aligner and language backbone are updated, preserving visual representations and general reasoning (\papertabref{tab:general}).

\begin{table}[!htb]
\caption{\textbf{Post-training hyperparameters for \modelname{}.} SFT co-trains agent-synthesized trajectories with general multimodal and spatial QA data self-generated by Qwen3.8-27B (zero external robot data); GRPO RL runs purely inside synthesized SAPIEN environments.}
\label{tab:hparams}
\centering
\footnotesize
\setlength{\tabcolsep}{4pt}
\renewcommand{\arraystretch}{1.15}
\begin{tabularx}{\linewidth}{@{}l>{\raggedright\arraybackslash}X>{\raggedright\arraybackslash}X@{}}
\toprule
Component & SFT & GRPO RL\\
\midrule
Backbone / init & Qwen3.8-27B (vision encoder frozen) & SFT checkpoint (frozen SFT KL reference)\\
Data / environments & 6{,}835 agent-solved trajectories + recovery forks + Qwen3.8-27B self-generated QA & 5{,}000 synthesized SAPIEN environments\\
Precision / context & bfloat16 / 24{,}576 tokens & bfloat16 / 4{,}096 max output tokens per turn\\
Batch / rollouts & Global batch size 64 & 176 groups (128 + 48 oversampling) $\times$ 16 = 2{,}816 rollouts/step\\
Compute / schedule & Peak LR $1.4\times 10^{-5}$, 5\% warm-up, cosine decay & 160 GPUs (20 eight-GPU nodes), 20 update steps\\
Objective / reg. & Cross-entropy on verified trajectories & Binary 0/1 \texttt{check()} reward, \texttt{low\_var\_kl}\\
Episode horizon & Median 8 turns (up to 4 actions/turn) & Max 40 turns (up to 4 actions/turn)\\
\bottomrule
\end{tabularx}
\end{table}

\paragraph{Environment scaling protocol.}
To isolate the effect of synthesized environment diversity (\paperfigref{fig:scaling}), we train checkpoints on nested pools of 100, 500, 1{,}000, and 5{,}000 synthesized SAPIEN environments using the same post-training pipeline and matched training time at every scale, and evaluate each checkpoint across all 3{,}000 LIBERO-Pro rollouts. \papertabref{tab:scaling-detail} reports the exact solved counts per suite together with mean turns on solved episodes across all four scales.

\begin{table}[!htb]
\caption{\textbf{Exact 3{,}000-rollout LIBERO-Pro scaling counts across pool sizes.} Each suite comprises 1{,}000 rollouts (500 position and 500 task perturbations). Mean turns are computed over solved episodes.}
\label{tab:scaling-detail}
\centering
\footnotesize
\setlength{\tabcolsep}{3pt}
\renewcommand{\arraystretch}{1.15}
\begin{tabular*}{\linewidth}{@{\extracolsep{\fill}}lccccc@{}}
\toprule
Training pool & Spatial (/1{,}000) & Object (/1{,}000) & Goal (/1{,}000) & Overall (/3{,}000) & Solved turns\\
\midrule
100 environments & 88 (8.8\%) & 148 (14.8\%) & 96 (9.6\%) & 332 (11.07\%) & 16.37\\
500 environments & 240 (24.0\%) & 219 (21.9\%) & 176 (17.6\%) & 635 (21.17\%) & 15.55\\
1{,}000 environments & 273 (27.3\%) & 344 (34.4\%) & 219 (21.9\%) & 836 (27.87\%) & 13.29\\
\midrule
\textbf{5{,}000 environments (\modelname{})} & \textbf{708 (70.8\%)} & \textbf{534 (53.4\%)} & \textbf{440 (44.0\%)} & \textbf{1{,}682 (56.07\%)} & \textbf{11.29}\\
\bottomrule
\end{tabular*}
\end{table}

\section{Evaluation protocols and baseline configurations}
\label{app:eval}

\paragraph{LIBERO-Pro (3{,}000 rollouts).}
All evaluations follow the 6-cell LIBERO-Pro protocol of CaP-X~\citep{capx}, also adopted by Guava~\citep{guava}: three suites (Object, Goal, Spatial) crossed with two perturbation types (object-position and task-instruction), 10 tasks per cell, and all 50 official initial states per task (seeds 1--50, totalling 3{,}000 rollouts per model). \modelname{} is served in bfloat16 via vLLM~\citep{vllm} with temperature 0.3, top-$p$ 1.0, max 4{,}096 output tokens per turn, and max 40 turns per episode. Exact 3{,}000-episode solved counts for our evaluated models in \papertabref{tab:liberopro} are 1{,}682/3{,}000 for \modelname{} (56.07\%), 1{,}977/3{,}000 for PEAR-27B (65.90\%), 213/3{,}000 for RoboUniverse-35B-A3B (7.10\%), 98/3{,}000 for Qwen3.8-27B Base (3.27\%), and 23/3{,}000 for nanoAstra-9B (0.77\%). For fair cross-embodiment zero-shot evaluation, nanoAstra-9B receives camera/table-frame alignment with its unused arm masked, and RoboUniverse-35B-A3B uses public Franka kinematics.

\paragraph{RoboLab-120 (1{,}200 episodes).}
We evaluate \modelname{} zero-shot in Isaac Sim across all 120 RoboLab tasks (64 Simple, 39 Moderate, 17 Complex) with 10 rollouts per task (1{,}200 episodes total; 472/1{,}200 = 39.3\% overall) on a Franka arm equipped with an unseen Robotiq parallel-jaw gripper, reporting both binary completion rate and RoboLab's graded subtask Score (53.0). PEAR-27B and DROID-trained leaderboard baselines in \papertabref{tab:robolab} follow Table~5 of \citet{pear}.

\paragraph{Real-world PiPER + Pika (30 trials and dim lighting).}
Real-world experiments use an AgileX PiPER 6-DoF arm with a Pika parallel-jaw gripper, a calibrated fixed front RGB camera, and a calibrated wrist RGB camera. Each of the 10 tasks in \papertabref{tab:real} is evaluated over 3 consecutive trials (30 trials total; 23/30 = 76.7\%), where a trial succeeds only if the target physical relation is stably achieved after gripper release and retreat. The dim-lighting rollout (\paperfigref{fig:dim}) is evaluated separately on \emph{Ball on can} (Hard) with ambient lighting reduced so front and wrist pixel intensities drop to about 20\% of the bright-light frames, using the identical instruction and camera calibration.

\paragraph{Inference latency benchmark (60 unique tasks).}
All models in \papertabref{tab:latency} are benchmarked on identical $2\times$H100 GPUs with tensor parallelism TP=2 and strict single-stream batch size BS=1 (concurrency 1 per TP=2 replica) across all 60 unique LIBERO-Pro tasks (6 cells $\times$ 10 tasks, seed~1). We log pure LLM call latency (\emph{Turn LLM} mean and P90), prompt and generated token counts, effective output rate (\modelname{} 79.5 vs.\ PEAR-27B 77.1\,tok/s), simulator and tool overhead per episode (\modelname{} 24.3\,s, PEAR-27B 16.2\,s, RoboUniverse-35B-A3B 125.1\,s, nanoAstra-9B 75.7\,s), and end-to-end episode latency across both all 60 episodes and solved episodes (\modelname{} 58.8\,s across 34/60 solved vs.\ PEAR-27B 177.6\,s across 37/60 solved).

\section{General benchmarks}
\label{app:general}

\papertabref{tab:general-full} reports the complete paired evaluation of \modelname{} against its base model Qwen3.8-27B~\citep{qwen38} across the seven general benchmarks of \papertabref{tab:general} under identical serving and sampling settings (\texttt{evalscope}~1.13.0~\citep{evalscope}, BF16 vLLM~\citep{vllm}, thinking-mode sampling with temperature $1.0$, top-$p$ $0.95$, and top-$k$ $20$). RealWorldQA, MMMU, MMMU-Pro, ERQA, and IFBench use an 81{,}920-token generation limit (5{,}825 total generations per model), while MATH-500 uses a 32{,}768-token limit. Confidence intervals are computed via 2{,}000-resample paired bootstrap over questions.

Across all seven benchmarks, every paired 95\% confidence interval straddles zero.

\begin{table}[!htb]
\caption{\textbf{Paired general evaluation of \modelname{} against Qwen3.8-27B.} Both models are evaluated on the exact same questions with \texttt{evalscope}~1.13.0 and BF16 vLLM ($T=1.0$, top-$p=0.95$, top-$k=20$; 81{,}920 max tokens except MATH-500 at 32{,}768). Intervals are 95\% paired bootstrap confidence intervals over questions (2{,}000 resamples); changes are computed before rounding.}
\label{tab:general-full}
\centering
\footnotesize
\setlength{\tabcolsep}{3.5pt}
\renewcommand{\arraystretch}{1.15}
\begin{tabular*}{\linewidth}{@{\extracolsep{\fill}}lllrrrr@{}}
\toprule
Benchmark & Category & Questions & Base & \modelname{} & Change & 95\% CI\\
\midrule
MATH-500~\citep{math500} & math & 500 & 98.0 & 98.6 & $+0.6$ & $[-0.4,\,+1.8]$\\
RealWorldQA~\citep{realworldqa} & visual QA & 765 & 83.4 & 83.5 & $+0.1$ & $[-2.1,\,+2.5]$\\
MMMU (val)~\citep{mmmu} & multimodal & 900 & 79.9 & 79.6 & $-0.3$ & $[-2.2,\,+1.4]$\\
MMMU-Pro (standard)~\citep{mmmupro} & multimodal & 1{,}730 (10-opt) & 76.3 & 75.4 & $-0.9$ & $[-2.4,\,+0.8]$\\
MMMU-Pro (vision) & multimodal & 1{,}730 & 74.9 & 73.9 & $-1.0$ & $[-2.4,\,+0.4]$\\
IFBench~\citep{ifbench} & instruction following & 300 (strict) & 73.3 & 71.3 & $-2.0$ & $[-6.3,\,+2.7]$\\
ERQA~\citep{erqa} & embodied reasoning & 400 & 65.0 & 62.8 & $-2.3$ & $[-6.3,\,+1.8]$\\
\bottomrule
\end{tabular*}
\end{table}

\end{document}